\documentclass[11pt]{article}

\usepackage[english]{babel}

\usepackage{tabularx}
\usepackage{booktabs} 

\usepackage[utf8]{inputenc} 
\usepackage[T1]{fontenc}    
\usepackage{hyperref}       
\usepackage{url}            
\usepackage{booktabs}       
\usepackage{nicefrac}       
\usepackage{microtype}      
\usepackage{xcolor}         

\newcommand{\blue}{\color{black}}

\usepackage{comment}
\usepackage{algorithm}
\usepackage{algorithmic}
\usepackage{multicol}
\usepackage{graphicx}
\usepackage{subfigure}
\usepackage{titletoc}

\usepackage{multirow}
\usepackage{caption}
\usepackage{amsmath}
\usepackage{mathtools}
\usepackage{amsthm}
\usepackage{makecell}
\usepackage{amssymb}
\usepackage{array}

\newcolumntype{L}[1]{>{\raggedright\let\newline\\\arraybackslash\hspace{0pt}}m{#1}}
\newcolumntype{C}[1]{>{\centering\let\newline\\\arraybackslash\hspace{0pt}}m{#1}}
\newcolumntype{R}[1]{>{\raggedleft\let\newline\\\arraybackslash\hspace{0pt}}m{#1}}

\usepackage{pifont}
\newcommand{\cmark}{\ding{51}}%
\newcommand{\xmark}{\ding{55}}%

\usepackage{graphicx}
\usepackage{float} 
\usepackage[capitalize,noabbrev]{cleveref}

\DeclareMathOperator*{\argmin}{arg\,min}
\newcommand{\argminF}{\mathop{\mathrm{argmin}}\limits}

\theoremstyle{plain}

\theoremstyle{definition}

\theoremstyle{remark}

\usepackage[flushleft]{threeparttable}
\usepackage{booktabs,amsmath,siunitx} 

\usepackage[letterpaper,top=2cm,bottom=2cm,left=3cm,right=3cm,marginparwidth=1.75cm]{geometry}

\usepackage{amsmath}
\usepackage{graphicx}

\usepackage{tikz}
\newcommand\copyrightnotice[1]{
    \begin{tikzpicture}[remember picture,overlay]
    \node[anchor=south,yshift=35pt] at (current page.south) {\fbox{\parbox{\dimexpr\textwidth-\fboxsep-\fboxrule\relax}{#1}}};
    \end{tikzpicture}
}

\begin{document}

\title{Vertical Federated Learning: Concepts, Advances and Challenges}
\author{Yang Liu$^1$\footnote{corresponding author, liuy03@air.tsinghua.edu.cn.}, Yan Kang$^2$, Tianyuan Zou$^1$, Yanhong Pu$^1$, Yuanqin He$^2$, Xiaozhou Ye$^3$, \\Ye Ouyang$^3$,
Ya-Qin Zhang$^1$ and Qiang Yang$^{2,4}$}
\date{
\textit{\small $^1$ Institute for AI Industry Research, Tsinghua University, Beijing, China. \\
$^2$ Webank, Shenzhen, China. \\
$^3$ AsiaInfo Technologies, Beijing, China.\\
$^4$ Hong Kong University of Science and Technology, Hong Kong, China. \\
}}
\maketitle

\copyrightnotice{\tiny This work has been submitted to the IEEE for possible publication. Copyright may be transferred without notice, after which this version may no longer be accessible.}

\begin{abstract}
Vertical Federated Learning (VFL) is a federated learning setting where multiple parties with different features about the same set of users jointly train machine learning models without exposing their raw data or model parameters. Motivated by the rapid growth in VFL research and real-world applications, we provide a comprehensive review of the concept and algorithms of VFL, as well as current advances and challenges in various aspects, including effectiveness, efficiency, and privacy. We provide an exhaustive categorization for VFL settings and privacy-preserving protocols and comprehensively analyze the privacy attacks and defense strategies for each protocol. We propose a unified framework, termed VFLow, which considers the VFL problem under communication, computation, privacy, as well as effectiveness and fairness constraints. Finally, we review the most recent advances in industrial applications, highlighting open challenges and future directions for VFL. 
\end{abstract}

\section{Introduction}

\label{sec:introduction}

\begin{figure*}[!tb]
 \centering
  \subfigure[Horizontal Federated Learning]{
 \begin{minipage}[t]{0.30\linewidth}
 \centering
 \includegraphics[width=1\linewidth]{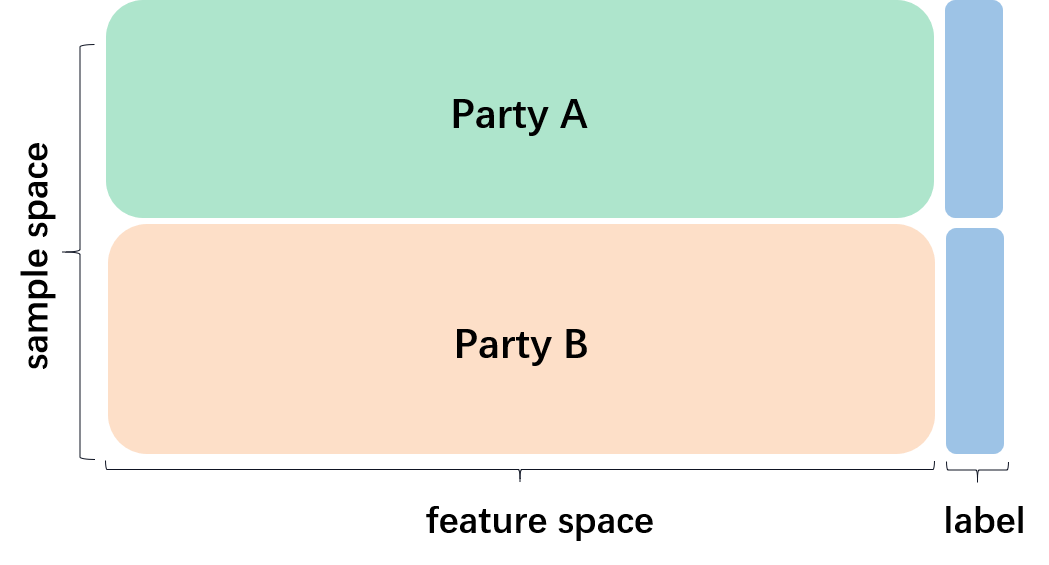}
 \label{fig:sub_fig_fl_setting_hfl}
 \end{minipage}
 }
 \subfigure[Vertical Federated Learning]{
 \begin{minipage}[t]{0.30\linewidth}
 \centering
 \includegraphics[width=1\linewidth]{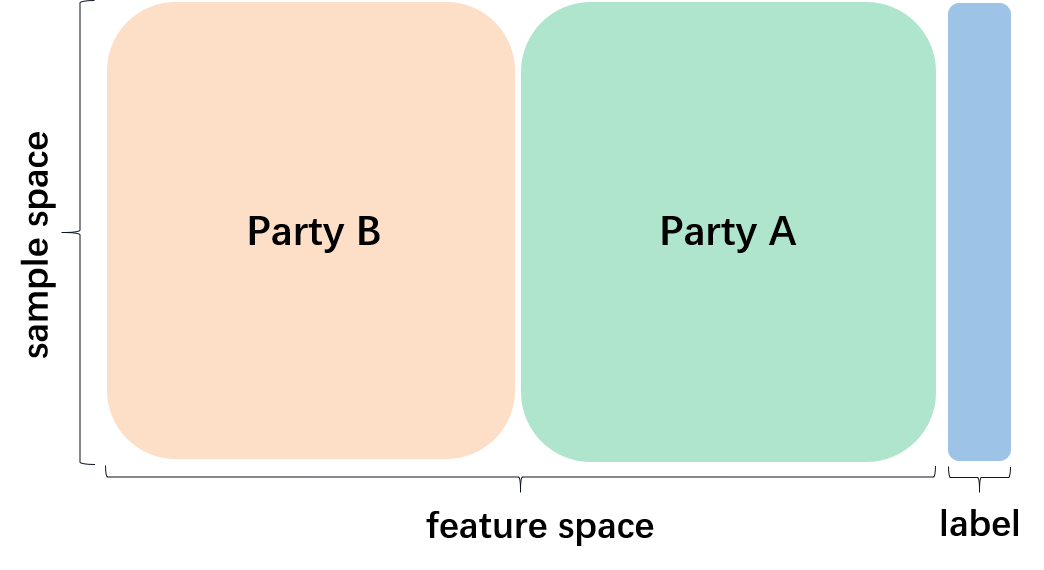}
 \label{fig:sub_fig_fl_setting_vfl}
 \end{minipage}
 }
 \subfigure[Federated Transfer Learning]{
 \begin{minipage}[t]{0.30\linewidth}
 \centering
 \includegraphics[width=1\linewidth]{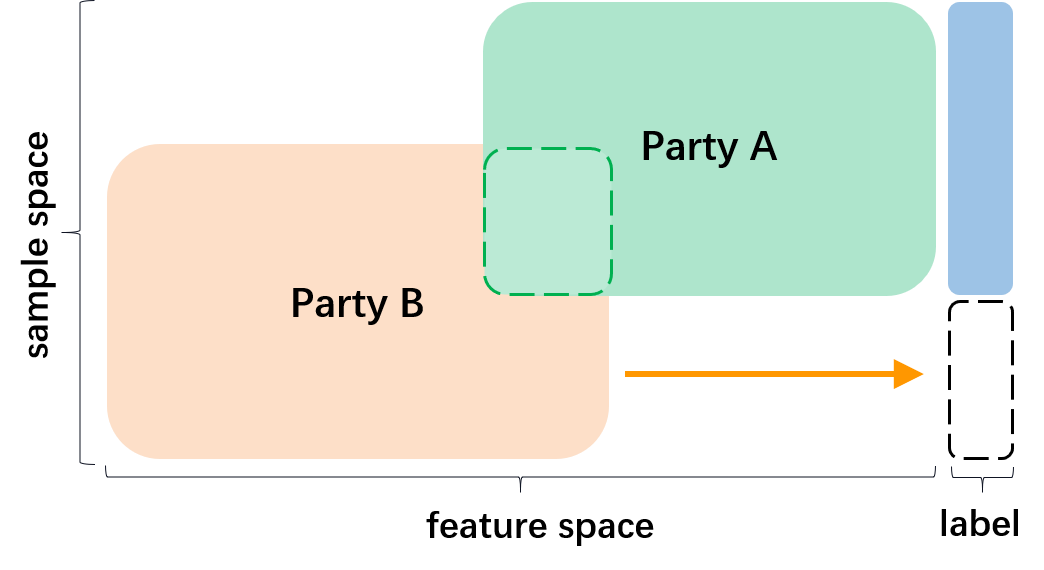}
 \label{fig:sub_fig_fl_setting_ftl}
 \end{minipage}
 }
 \caption{Three categories of Federated Learning
 }
 \label{fig:fl_settings}
\end{figure*}

Federated Learning (FL) \cite{yang2019federatedbook} is a novel machine learning paradigm where multiple parties collaboratively build machine learning models without centralizing their data. The concept of FL was first proposed by Google in 2016 \cite{McMahanMRA16} to describe a cross-device scenario where millions of mobile devices are coordinated by a central server while local data are not transferred. This concept is soon extended to a cross-silo collaboration scenario among organizations \cite{Yang2019FLconcept}, where a small number of reliable organizations join a federation to train a machine learning model. In \cite{Yang2019FLconcept}, FL is, for the first time, categorized into three categories based on how data is partitioned in the sample and feature space: Horizontal Federated Learning (HFL), Vertical Federated Learning (VFL) and Federated Transfer Learning (FTL) (See Figure \ref{fig:fl_settings}).

 \begin{itemize}
     \item HFL refers to the FL setting where participants share the same feature space while holding different samples. For example, Google uses HFL to allow mobile phone users to use their dataset to collaboratively train a next-word prediction model \cite{McMahanMRA16}.
     \item VFL refers to the FL setting where datasets share the same samples/users while holding different features. For example, Webank uses VFL to collaborate with an invoice agency to build financial risk models for their enterprise customers \cite{Cheng2020CACM}.
     \item FTL refers to the FL setting where datasets differ in both feature and sample spaces with limited overlaps. For example, EEG data from multiple subjects with heterogeneous distributions collaboratively build BCI models using FTL \cite{ju2020FTLEEG}.
 \end{itemize}


Due to their differences in data partitions, HFL and VFL adopt very different training protocols. Each party in HFL trains a local model and exchanges model updates (i.e., parameters or gradients) with a server, which aggregates the updates and sends the aggregating result back to each party. While in VFL, each party keeps both its data and model local but exchanges intermediate computed results. The output of the HFL training procedure is a global model shared among all parties, while each party in the VFL owns a separate local model after training. During inference time, each party in HFL uses the global model separately, while parties in VFL need to collaborate to make inferences. FL can also be categorized into "cross-device" and "cross-silo" settings~\cite{google_workshop_2019}. The cross-device FL may involve a vast number of mobiles or edge devices as the participating parties. In contrast, the participating parties in the cross-silo FL are typically a limited number of organizations. HFL can be either cross-device or cross-silo FL, while VFL typically belongs to the cross-silo FL. We compare these main differences between HFL, VFL, and FTL in Table \ref{tab:vfl_vs_hfl}. Note that Table \ref{tab:vfl_vs_hfl} compares the conventional cases of HFL, VFL, and FTL. As this research area experiences explosive growth, some special cases may deviate from Table \ref{tab:vfl_vs_hfl}.
 
\begin{table*}[!h]
 	\centering
	\small
	\caption{Comparison of main characteristics between conventional HFL, VFL and FTL.} 
	\begin{tabular}{C{3.4cm}||C{3.5cm}|C{3.4cm}|C{3.4cm}}
	    \hline
		 & HFL & \textbf{VFL} & FTL \\
	    \hline
	    \hline
		Data is different in & Sample space & Feature space &  Both\\
		\hline
	   	Scenarios & Cross-device/ Cross-silo & Cross-silo  & Mostly Cross-silo \\
		\hline
		What is exchanged?& Model parameters & Intermediate results & Intermediate results \\
		\hline
		What is kept local? & Local data &  Local data and model & Local data and model\\
		\hline
		Each party obtains  & A shared global model & A local model & A local model \\
		\hline
		Collaborative Inference? & No & Yes  & No \\
		\hline
	\end{tabular}
\label{tab:vfl_vs_hfl}
\end{table*}

The need for VFL has arisen and grown strongly in the industry in recent years. Companies and institutions owning only small and fragmented data have constantly been looking for compensating data partners to collaboratively develop artificial intelligence (AI) technology for maximizing data utilization \cite{li2021surveyTKDE,li2020applicationsurvey}. At the same time, data privacy and security regulations have been strengthened worldwide due to growing public concerns over data leakage and privacy breaches. Accordingly, many privacy-preserving projects and platforms supporting VFL have been developed in the past two years \cite{liu2021fate,romanini2021pyvertical,Fedlearner,he2020fedml,fedtree2022}, and the number of commercialized projects as well as the economic values of VFL have grown significantly. Since in VFL, data parties with different attributes of people are typically from different industrial segments, for example, a local bank and a local retailer, they are prone to collaborate rather than compete. 

{\blue{
While the applications and research on VFL have grown dramatically in recent years, there lacks a comprehensive survey on the advances, challenges, and potential research directions of VFL. Existing FL surveys focus either on HFL~\cite{google_workshop_2019,li2021survey,tan2022towards} or a limited perspective of VFL~\cite{yang2019federated,tran2022privacy}. } }
Therefore, we provide a comprehensive overview of current progress in VFL. We propose an exhaustive categorization for VFL settings and privacy-preserving protocols and discuss possible routes for improving effectiveness, efficiency, and privacy. In the end, we propose a unified framework, termed VFLow, which is extended from the original VFL definition and takes into account communication, computation, effectiveness, privacy, and fairness constraints. 

\begin{figure}[!hb]
 \centering
 \includegraphics[width=0.95\linewidth]{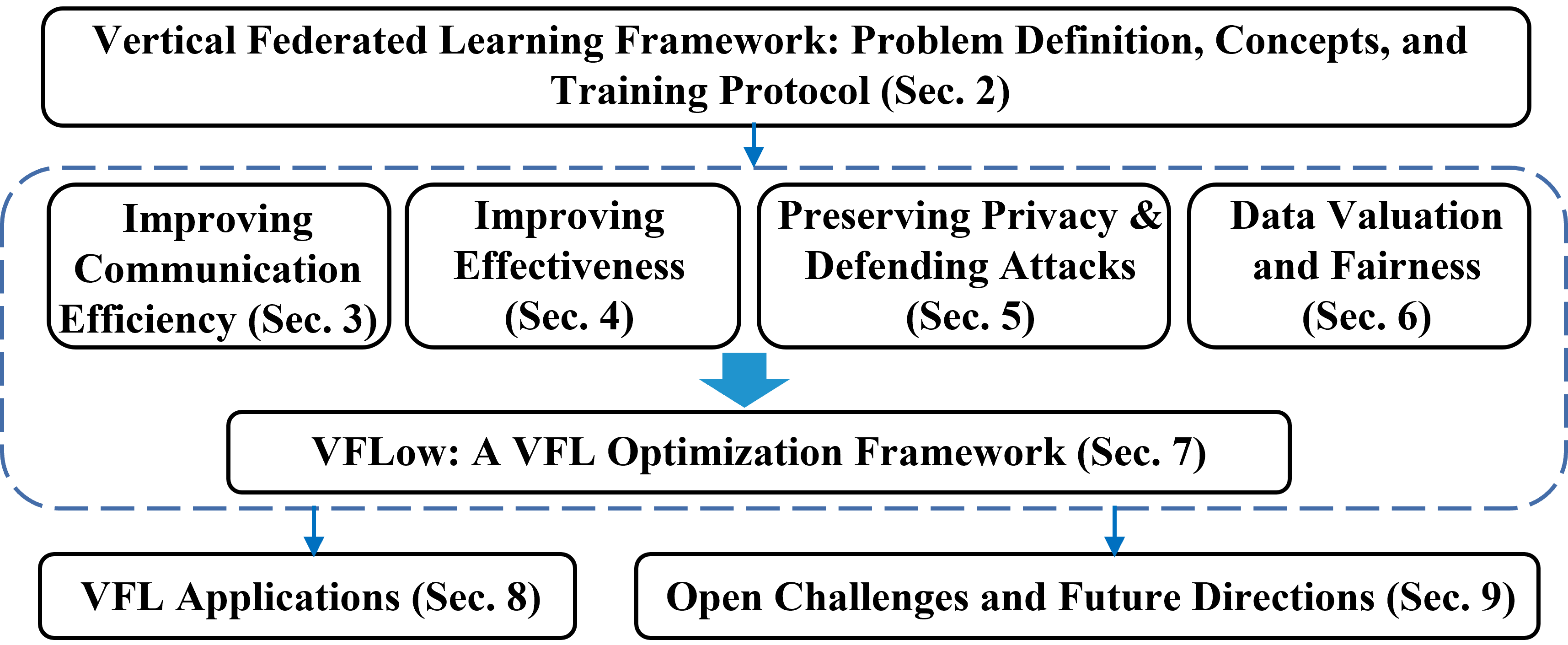}
 \caption{Relationships between sections in this work.}
 \label{fig:outline}
\end{figure}


This paper is organized as follows: Sec. \ref{overview} overviews VFL's concepts and training procedures. Building on Sec. \ref{overview}, Sec. \ref{efficiency}, Sec. \ref{effective}, and Sec. \ref{privacy} discuss the efficiency, effectiveness, privacy, and security aspects of VFL algorithms. Sec. \ref{fairness} discusses the challenges of data valuation, explainability, and fairness towards building a VFL ecosystem. {\blue{Sec. \ref{vflow} introduces \textbf{VFLow}, a VFL optimization framework guiding the design and optimization of VFL algorithms}}, and Sec. \ref{applied} discusses application-oriented algorithms built on VFL. Finally, Sec. \ref{challenge} discusses open challenges and future directions. {\blue{Figure \ref{fig:outline} dictates the relationships between sections in this work.}}

\section{Vertical Federated Learning framework}\label{overview}

In this section, we provide an overview of VFL formulation, algorithm, and variants.

\subsection{Problem Definition}

A VFL system aims to collaboratively train a joint machine learning (ML) model using a dataset $\mathcal{D} \triangleq \{(\mathbf{x}_i,y_i)\}_{i=1}^{N}$ with $N$ samples while preserving the privacy and safety of local data and models. We formulate the loss of VFL as follows.
\begin{equation}\label{eq:problem}
\min_{\mathbf \Theta} \ell(\mathbf \Theta; \mathcal{D}) \triangleq  \frac{1}{N}\sum^N_{i=1} f(\mathbf \Theta; \mathbf{x}_i,y_i)+ \lambda\sum_{k=1}^K\gamma(\mathbf \Theta)
\end{equation}
where $\mathbf \Theta$ denote the joint ML model; $f(\cdot)$ and $\gamma(\cdot)$ denote the loss function and regularizer and $\lambda$ is the hyperparatemer that controls the strength of $\gamma$. 

VFL assumes that data are partitioned by feature space. Following \cite{Yang2019FLconcept,liu2022fedbcd}, each feature vector $\mathbf{x}_i \in \mathbb{R}^{1 \times d}$ in $\mathcal{D}$ is distributed among $K$ parties $\{\mathbf{x}_{i,k} \in \mathbb{R}^{{1 \times d_k}}\}_{k=1}^{K}$, where $d_k$ is the feature dimension of party $k$, for $k\in [K-1]$, and the $K^{th}$ party has the label information $y_{i}=y_{i,K}$.  We refer to the $K^{th}$ party who owns the labels as \textit{active party} while the rest of parties as \textit{passive parties}. Each passive party $k$ has dataset $\mathcal{D}_k\triangleq \{\mathbf{x}_{i,k}\}_{i=1}^{N}$, while the active party has dataset $\mathcal{D}_K\triangleq \{\mathbf{x}_{i,K},y_{i,K}\}_{i=1}^{N}$.

Without loss of generality, we decompose $\mathbf \Theta$ into local models $\mathcal{G}_k$ parameterized by $\theta_k$, $k \in \{1,\cdots, K\}$, which operates only on local data, and a global module $\mathcal{F}_K$ parameterized by $\psi_K$, which is only accessible by the active party $K$. We rewrite the loss $f(\mathbf \Theta; \mathbf{x}_i,y_i)$ as:
\begin{equation}\label{eq:loss}
\begin{split}
\hspace{-0.05cm} & f(\mathbf \Theta; \mathbf{x}_i,y_i) \\ &= \mathcal{L}\left(\mathcal{F}_K\left(\psi_K;\mathcal{G}_1(\mathbf{x}_{i,1},\theta_1),...,\mathcal{G}_K(\mathbf{x}_{i,K},\theta_K)\right),y_{i,K}\right)
\end{split}
\end{equation}
where $\mathcal{L}$ denotes the task loss (e.g., {\blue{mean squared error
loss, cross-entropy loss, and hinge loss}}).

\begin{figure}[!tb]
 \centering
 \includegraphics[width=0.99\linewidth]{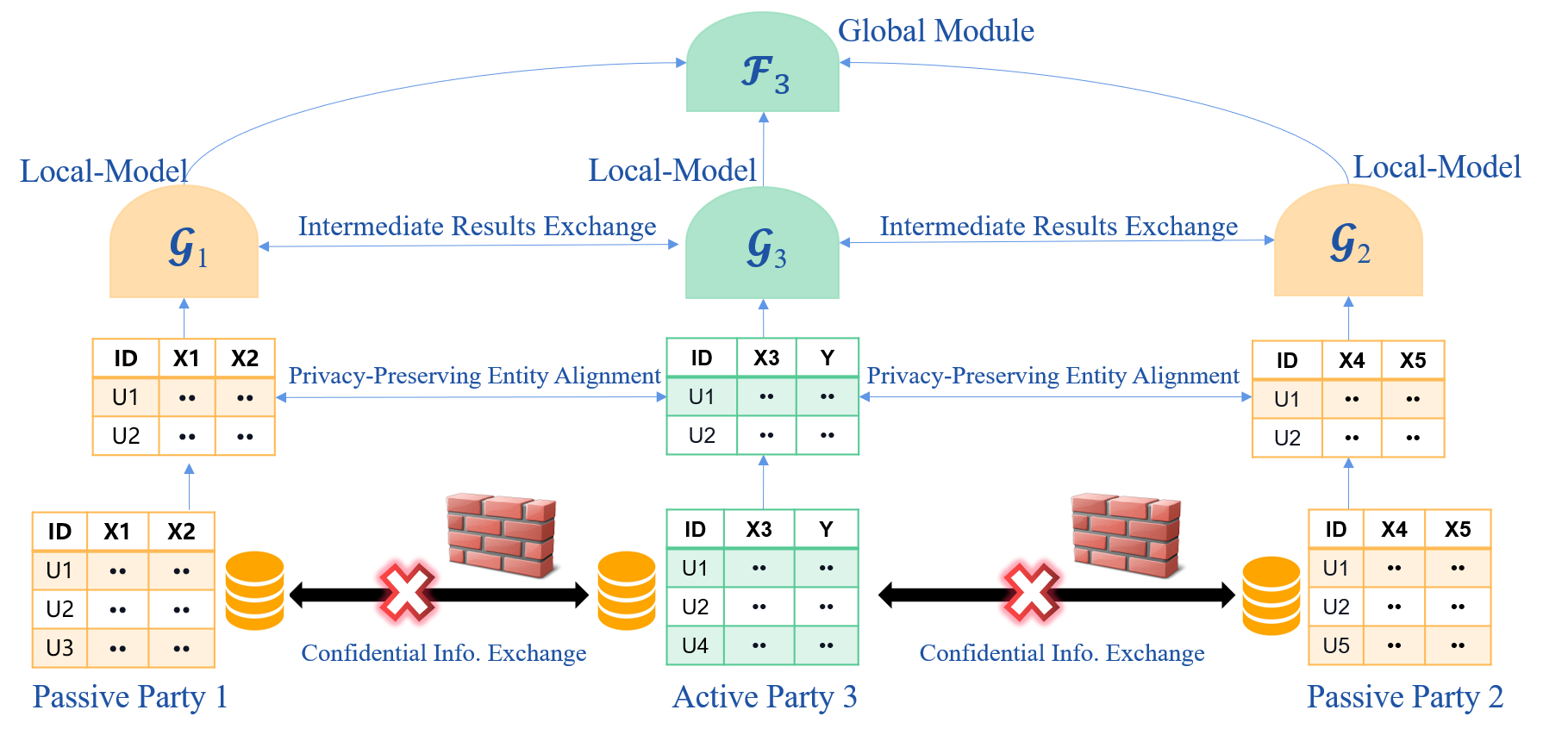}
 \caption{Illustration of the VFL system with three parties (two passive parties and one active party). {\blue{$\mathcal{G}_1, \mathcal{G}_2$, and $\mathcal{G}_3$ denote the local models of the three parties, respectively, and $\mathcal{F}_3$ denotes the global module owned by the active party. The VFL training protocol typically involves two steps: 1) the three parties align their samples via private entity alignment; 2) the three parties collaboratively train $\mathcal{G}_1, \mathcal{G}_2, \mathcal{G}_3$ and $\mathcal{F}_3$ in a privacy-preserving manner (see Section \ref{sec:vfl_training} for details).}}} 
 \label{fig:VFL_system}
\end{figure}
 
\cref{fig:VFL_system} pictorially overviews the architecture and core components of a VFL system.  
{\blue{Each party's local data are not exchanged during the collaboration.}} The local model $\mathcal{G}_k$ can take various forms including tree~\cite{SecureBoost}, linear and logistic regression (LR)~\cite{Yang2019FLconcept,liu2022fedbcd,gratton2018distributed,Gascon2016SecureLR,Ashish2004Linear,chen2021homomorphic}, support vector machine~\cite{yu2006privacy,xu2021fedv}, neural network (NN)~\cite{liu2020secure,Hu2019FDMLAC,fu2021label}, as well as K-means~\cite{li2022vflkmean} and EM algorithm ~\cite{ou2020homomorphicEM} etc. Although most of the existing VFL works consider linearly separable local models, recent works \cite{Dang2020kernel} also proposed kernel methods for incorporating non-linear learning over distributed features. 

\begin{figure}[!tb]
 \centering
 \subfigure[The global module is trainable and the active party has features (\textbf{splitVFL})]{
 \begin{minipage}[t]{0.45\linewidth}
 \centering
 \includegraphics[width=1\linewidth]{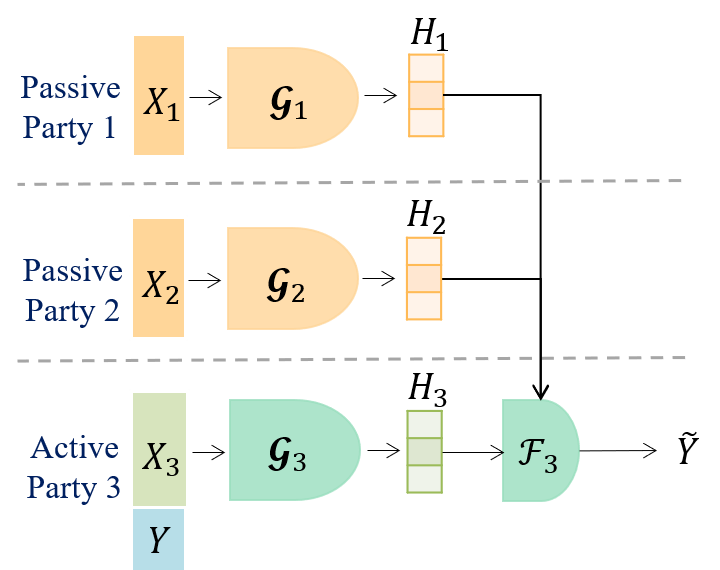}
 \label{fig:sub_fig_vfl_setting_trainable}
 \end{minipage}
 }
  \hspace{0.5em}
  \subfigure[The global module is non-trainable and the active party has features (\textbf{aggVFL}).]{
 \begin{minipage}[t]{0.45\linewidth}
 \centering
 \includegraphics[width=1\linewidth]{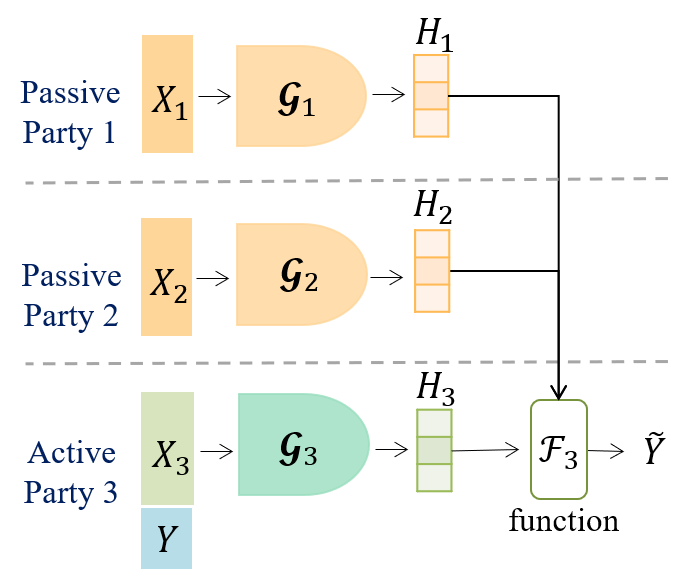}
 \label{fig:sub_fig_vfl_setting_nontrainable}
 \end{minipage}
 }
 \subfigure[The global module is trainable and the active party has no feature ($\textbf{splitVFL}_{c}$)]{
 \begin{minipage}[t]{0.45\linewidth}
 \centering
 \includegraphics[width=1\linewidth]{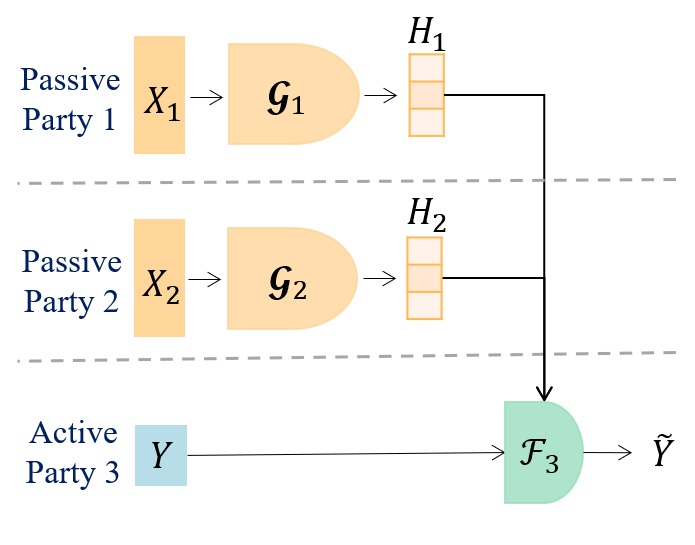}
 \label{fig:sub_fig_vfl_setting_splitNN}
 \end{minipage}
 }
  \hspace{0.5em}
 \subfigure[The global module is non-trainable and the active party has no feature ($\textbf{aggVFL}_{c}$)]{
 \begin{minipage}[t]{0.45\linewidth}
 \centering
 \includegraphics[width=1\linewidth]{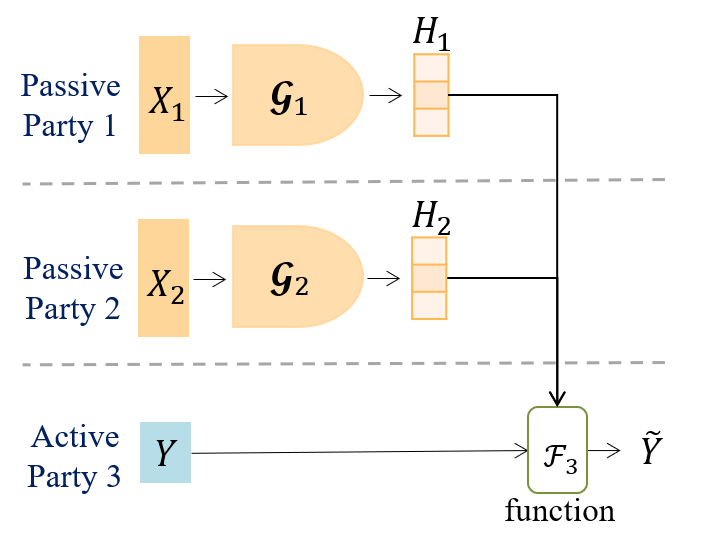}
 \label{fig:sub_fig_vfl_setting_idiotServer}
 \end{minipage}
 }
\caption{Four major variants of VFL illustrated with one active party and two passive parties.}
\label{fig:vfl_setting_variants}
\end{figure}

The global module $\mathcal{F}_K$ can be either \textit{trainable} \cite{fu2021label,jin2021cafe,liu2021rvfr} or \textit{non-trainable}\cite{fu2021label,zou2022defending}. If a \textit{trainable} global module is in place, this VFL scenario is coincident with the vertical splitNN~\cite{VerticalSplitNN}, where the whole model is splitted into different parties, thus we term it \textbf{splitVFL} (see \cref{fig:sub_fig_vfl_setting_trainable}). If the global module is \textit{non-trainable}, it serves as an aggregation function, such as Sigmoid (for NN) or an optimal split finding function (for tree), that aggregates parties' intermediate results. We term this scenario \textbf{aggVFL} (see \cref{fig:sub_fig_vfl_setting_nontrainable}). 


Another variant of VFL is when the active party has no features and thus it provides no local model. In this variant the active party plays the role of a \textbf{c}entral server. We refer to the active party providing no feaures in splitVFL and aggVFL, respectively, as \textbf{splitVFL$_c$} and \textbf{aggVFL$_c$}.
We illustrate these VFL variants in Figure \ref{fig:vfl_setting_variants} and summarize their architectural differences in Table \ref{tab:3-vfl-variants}.


\begin{table}[!htb]
	\centering
\small
	\caption{Comparison of splitVFL and aggVFL}
	\begin{tabular}{c||c|c}
	     \hline
		 & splitVFL &  aggVFL  \\
        \hline
        \hline
        \multirow{1}*{\shortstack{Has passive party models $\mathcal{G}_i, i=1,...,K$-1?}} & \multirow{1}*{Yes} & \multirow{1}*{Yes}  \\
        \hline
		\multirow{1}*{\shortstack{Is global module $\mathcal{F}_K$ trainable?}} & \multirow{1}*{Yes}  & \multirow{1}*{No}\\
        \hline
		\multirow{1}*{\shortstack{If active party $K$ has no features}} & \multirow{1}*{splitVFL$_c$} & \multirow{1}*{aggVFL$_c$}\\ 
		\hline
	   
	\end{tabular}
\label{tab:3-vfl-variants}
\end{table}

{\blue{In a typical VFL system, passive parties communicate only with the active party, which serves as the coordinator that orchestrates the training and inference procedures. In some scenarios, a third party is involved and responsible for encryption and decryption~\cite{liu2022fedbcd}. }}


\subsection{VFL Training Protocol}\label{sec:vfl_training}
  
 
In this section, we describe a general training protocol for VFL, which consists of two steps: $1)$ Entity Alignment; $2)$Privacy-preserving training. See Figure \ref{fig:VFL_system}.

{\blue{\textbf{Privacy-Preserving Entity Alignment.}}} The very first step for a VFL system to start a collaborative training process is to align the data used for the training. This process can be referred to as entity alignment, which adopts private set intersection techniques to find the common sample IDs without revealing unaligned dataset. We discuss these techniques in Sec. \ref{privacy}.  
{\blue{Whereas conventional VFL frameworks mostly consider entity alignment with exact IDs, recent studies \cite{NEURIPS2022_fuzzy_align} also demonstrated a coupled design for fuzzy identifiers to enable one-to-many alignment, which  could be an interesting future direction of VFL.}}


{\blue{\textbf{Privacy-Preserving Training by Exchanging Intermediate Results.}}} After the alignment, participating parties can start training the VFL model using the aligned samples. 
The most common training protocol is using gradient descent \cite{Li2007PPGD}, which requires parties to transmit local model outputs and corresponding gradients, {\blue{together termed intermediate results,}} instead of local data. Algorithm \ref{alg:first_order_VFL_he} describes a general VFL training procedure based on neural networks using stochastic gradient descent (SGD). Specifically, each party $k$ computes its local model output $H_{k} = \mathcal{G}_i(\mathbf{x}_{k},\theta_k)$ on a mini-batch of samples $\mathbf{x}$ and sends $H_{k}$ to the active party. With all the $\{H_{k}\}_{k=1}^K$, the active party computes the training loss following Eq. (\ref{eq:problem}). 
Then, the active party computes the gradients $\frac{\partial \ell}{\partial \psi_K}$ of its global module and updates its global module using $\frac{\partial \ell}{\partial \psi_K}$. Next, the active party computes the gradients $\frac{\partial \ell}{\partial H_{k}}$ for each party and transmits them back. Finally, each party $k$ computes the gradient of its local model $\theta_k$ as follows:
 \begin{equation} \label{eq:local_gradient_computation}
     \nabla_{\theta_k} \ell = \frac{\partial \ell}{\partial \theta_k} = \sum_i\frac{\partial \ell}{\partial H_{i,k}} \frac{\partial H_{i,k}}{\partial \theta_k}
 \end{equation}
and updates its local model. This procedure iterates until convergence. 

To prevent privacy leakage from the intermediate results $H_{k}$ and gradients $\frac{\partial \ell}{\partial H_{k}}$, Crypto-based privacy-preserving techniques such as Homomorphic Encryption (HE) (denoted as $[[\cdot]]$), Secure Multi-Party Computation (MPC) and Trusted Execution Environment (TEE) can be introduced into the VFL protocol to protect the crucial information from inner and outside attackers. For example, instead of sending $H_{k}$, each party $k$ sends $[[H_{k}]]$ to the active party, who in turn sends $[[\frac{\partial \ell}{\partial H_{k}}]]$ back to each party. {\blue{A third-party collaborator is often responsible for encryption and decryption.}} Other privacy-preserving techniques, such as Differential  Privacy (DP) and Gradient Discretization (GD) can also be applied to enhance the privacy and security of the VFL system. We provide detailed comparisons of these techniques in Sec. \ref{privacy}.

\begin{algorithm}[!tb]
 \caption{A General VFL Training Procedure.}
 \textbf{Input}: learning rates $\eta_1$ and $\eta_2$\\
 \textbf{Output}: Model parameters $\theta_1$, $\theta_2$ ... $\theta_K$, $\psi_K$ \\
 \begin{algorithmic}[1]
 \STATE Party 1,2,\dots,$K$, initialize $\theta_1$, $\theta_2$, ... $\theta_K$, $\psi_K$. \\
 \FOR{each iteration $j=1,2, ...$}
 \STATE Randomly sample a mini-batch of samples $\mathbf{x}\subset\mathcal{D}$
 \FOR{each party $k$=1,2,\dots,$K$ in parallel}
 \STATE Party $k$ computes $H_{k} = \mathcal{G}_k(\mathbf{x}_{k},\theta_k);$
 \STATE Party $k$ sends $\{H_{k}\}$ to party $K$;
 \ENDFOR
 \STATE Active party $K$ updates $\psi^{j+1}_K = \psi^{j}_K - \eta_1 \frac{\partial \ell}{\partial \psi_K}$;
 \STATE Active party $K$ computes and sends $\frac{\partial \ell}{\partial H_{k}}$ to all other parties;
 \FOR{each party $k$=1,2,\dots,$K$ in parallel}
 \STATE Party $k$ computes $\nabla_{\theta_k} \ell$ with \cref{eq:local_gradient_computation};
 \STATE Party $k$ updates $\theta^{j+1}_k = \theta^{j}_k - \eta_2 \nabla_{\theta_k} \ell$;
 \ENDFOR
 \ENDFOR
 \end{algorithmic}
 \label{alg:first_order_VFL_he}
\end{algorithm}



\subsubsection{Tree-based VFL}
 
Tree-based VFL complies with the architecture depicted in Figure \ref{fig:VFL_system} and follows the general loss defined in Eq. (\ref{eq:loss}) for conducting VFL training, but it differs from the NN-based VFL in local models $\mathcal{G}_k, k \in \{1,...,K\}$, the global module $\mathcal{F}_K$ as well as the specific training process at each party.

In tree-based VFL, the local model $\mathcal{G}_k$ at each party $k$ consists of multiple partial tree models that each partial tree model, together with its counterparts from other parties, form a complete tree model. The $\mathcal{F}_K$ is an aggregation function that identifies the optimal feature split based on feature splitting information received from all parties. 

The literature has proposed various GBDT-based VFL algorithms~\cite{SecureBoost,chen2021secureboost_plus,feng2019securegbm,wen2021securexgb,tian2020federboost,li2022opboost,xie2022mpfedxgb}. 
SecureBoost~\cite{SecureBoost}, SecureBoost+~\cite{chen2021secureboost_plus}, and SecureGBM~\cite{feng2019securegbm} exploit additive homomorphic encryption (HE) to encrypt residual errors and feature histograms transmitted between active and passive parties. SecureXGB~\cite{wen2021securexgb} and Pivot~\cite{pivot2020} utilize secret sharing mixed with additive HE to encrypt transmitted information. FederBoost~\cite{tian2020federboost} and OpBoost~\cite{li2022opboost} adopt differential privacy to protect individual data trying to achieve a better balance between privacy and efficiency.

Random Forest~\cite{tin1995rdf} (RF) is another popular tree-based ensemble algorithm that has been integrated into VFL. RF-based VFL algorithms~\cite{liu2020federated,yao2022efficient,hou2021verifiable} typically leverage bagging and optimized parallelism to enhance the training and inference efficiency. Federated Forest~\cite{liu2020federated} introduces a third party and applies RSA encryption to protect data privacy. VFRF~\cite{yao2022efficient} adopts randomized iterative affine cipher (RIAC)~\cite{Kyle2018affine} to encrypt transmitted information. VPRF~\cite{hou2021verifiable}, a verifiable privacy-preserving random forest scheme, is proposed to verify data integrity and preserve data privacy.

\section{Improving Communication Efficiency}
\label{efficiency}

\begin{table*}[ht!]
\centering
\scriptsize
\caption{Summary of existing works that aim to improve the efficiency of VFL. In the Model column, the LR denotes logistic regression, NN denotes Neural Network, XGB denotes XGBoost and GBDT denotes gradient boosting decision tree. In the Convergence Rate column, $T$ represents the total number of local iterations and $\Delta$ represents stochastic variance.}
\begin{tabular}{m{1.9cm}<{\centering}||m{2.8cm}<{\centering}|m{1.4cm}<{\centering}|m{1.4cm}<{\centering}|m{1.4cm}<{\centering}|m{3.9cm}<{\centering}}
\hline
\multicolumn{1}{c||}{\shortstack{Category}} & \multicolumn{1}{c|}{\shortstack{Existing \\ Work}}  &
\multicolumn{1}{c|}{\shortstack{VFL \\ Setting}}  & \multicolumn{1}{c|}{\shortstack{Model}}   &
\multicolumn{1}{c|}{\shortstack{\tiny 
 Convergence \\ \tiny Rate}} & 
\multicolumn{1}{c}{\shortstack{Core \\ Method}} \\

\hline
\hline
\multirow{8}*{\scriptsize\shortstack{Multiple \\Client\\Updates}} & FedBCD~\cite{liu2022fedbcd} & splitVFL / aggVFL & LR/NN & $O(1/\sqrt{T})$ & \scriptsize Block coordinate descent w/ multiple local updates \\
\cline{2-6}
~ & Flex-VFL~\cite{Castiglia2022FlexibleVF} & $\text{splitVFL}_c$ & NN  & $O(1/\sqrt{T})$ & \scriptsize Customized \# of local updates constrained by time \\
\cline{2-6}
~ & AdaVFL~\cite{Jie2022adavfl} & $\text{aggVFL}_c$ & NN  & $O(1/\sqrt{T})$ & \scriptsize Customized \# of local updates through optimization  \\
\cline{2-6}
\\[-1em]
~ & VIMADMM~\cite{xie2022improving} & splitVFL / aggVFL & NN  & - & \scriptsize Alternative direction method of multipliers \\
\cline{2-6}
~ & CELU-VFL~\cite{fu2022communicationefficient} & splitVFL & NN & $O(\Delta/\sqrt{T})$ & \scriptsize Cache-based mechanism for local updates\\
\hline
\multirow{20}*{\scriptsize\shortstack{Asynchronous \\Coordination}} & GP-AVFL~\cite{li2020asynchronousVFL} & $\text{aggVFL}$ & LR/NN & - & \scriptsize Asynchronous training with gradient prediction \\
\cline{2-6}
~ & AVFL~\cite{cai2022avfl} & aggVFL & LR & - & \scriptsize Backup-based straggler-resilient scheme \\
\cline{2-6}
~ & T-VFL~\cite{zhang2022lowlatency} & $\text{splitVFL}_c$ & NN & $O(1/\sqrt{T})$ & \scriptsize Channel-aware user scheduling poicy\\
\cline{2-6}
~ & VAFL~\cite{chen2020vafl} & $\text{splitVFL}_c$ & LR/NN & $O(1/\sqrt{T})$ & \scriptsize Asynchronous query-response strategy \\
\cline{2-6}
~ & $\text{FDML}$~\cite{Hu2019FDMLAC} & $\text{aggVFL}{_c}$ & LR/NN & $O(1/\sqrt{T})$ & \scriptsize Asynchronous local updates w/ the same data order\\
\cline{2-6}
~ & $\text{AFAP}$~\cite{Gu2021Asynchronous} & aggVFL & LR & $O(e^{-T})$ & \scriptsize Tree-structured asynchronous communication (TSAC) \\
\cline{2-6}
~ & $\text{AsySQN}$~\cite{zhang2021asysqn} & aggVFL & LR & $O(e^{-T})$ & \scriptsize TSAC \& quasi-Newton method \\
\cline{2-6}
~ & $\text{VFB}^2$~\cite{zhang2021secure} & aggVFL & LR & $O(e^{-T})$ & \scriptsize TSAC \& bi-level parallel update\\
\cline{2-6}
~ & FDSKL~\cite{gu2020fdskl} & aggVFL & LR & $O(1/T)$ & \scriptsize TSAC \& random features \& doubly stochastic gradient \\
\cline{2-6}
~ & FedGBF~\cite{han2022fedgbf} & aggVFL & GBDT & - & \scriptsize Use RT as the base learner for learning GBDT \\
\cline{2-6}
~ & VF$^{2}$Boost\cite{fu2021vf2boost} & aggVFL & GBDT & - & \scriptsize Concurrent protocol \& customized Paillier HE \\
\hline
\multirow{6}*{\scriptsize\shortstack{One-Shot \\Communication}} & FedOnce~\cite{wu2022practical} & splitVFL & NN & - & \scriptsize Unsupervised learning by predicting noise \\
\cline{2-6}
~ & AE-VFL~\cite{cha2021implementing} & splitVFL & NN & - & \scriptsize Unsupervised learning using autoencoder\\
\cline{2-6}
\\[-1em]
~ & CE-VFL~\cite{khan2022communicationefficient} & splitVFL & NN & - & \scriptsize Unsupervised learning using PCA \& autoencoder
\\
\hline
\multirow{10}*{\scriptsize\shortstack{Compression}} & AVFL~\cite{cai2022avfl} & aggVFL & LR & - & \scriptsize Principle component analysis \\
\cline{2-6}
~ & CE-VFL~\cite{khan2022communicationefficient} & splitVFL & NN & - & \scriptsize Autoencoder and principle component analysis\\
\cline{2-6}
 & SecureBoost+~\cite{chen2021secureboost_plus} & aggVFL & XGB & - & \scriptsize Encode encrypted first-order and second-order gradients into a single message \\
\cline{2-5}
 & eHE-SecureBoost~\cite{xu2021efficient} & aggVFL & XGB & - & ~ \\
\cline{2-6}
~  &C-VFL~\cite{castiglia2022compressed} & splitVFL & NN & $O(1/\sqrt{T})$ & \scriptsize Arbitrary compression scheme \\
\cline{2-6}
~ & GP-AVFL+DESC~\cite{li2020asynchronousVFL} & $\text{aggVFL}$ & LR/NN & - & \scriptsize Double-end sparse compression\\

\hline
\\[-1em]
{\blue{\multirow{10}*{\scriptsize\shortstack{Sample \\ and Feature \\ Selection}}}} & Coreset-VFL~\cite{NEURIPS2022_coreset} & $\text{aggVFL}_c$ & \tiny LR, K-Mean & - & \scriptsize Coreset to select samples\\
\cline{2-6}
\\[-1em]
~ & FedSDG-FS~\cite{li2023fedsdg} & $\text{splitVFL}_c$ & NN & - & \scriptsize Stochastic dual-gate to select features \\
\cline{2-6}
\\[-1em]
~ & SFS-VFL~\cite{Zhang2022secfs} & $\text{aggVFL}_c$ & \tiny LR, KNN, SVM, GBT & - & \scriptsize Gini impurity to select features\\
\cline{2-6}
\\[-1em]
 & LESS-VFL\cite{castiglia2023less} & $\text{splitVFL}_c$ & NN & - & \scriptsize Group lasso regularization to select features \\
\cline{2-6}
\\[-1em]
 & FEAST~\cite{fu2023feast} & aggVFL & \tiny LR, SVM, XGB, NN  & - & \scriptsize Conditional mutual information to select features \\
\cline{2-6}
\\[-1em]
~  &VFLFS~\cite{feng2022vflfs} & splitVFL & NN & - &  \scriptsize Use trainable transformation matrix to select features \\

\hline
\end{tabular}
\label{tab:vfl_efficiency}
\end{table*}


In production VFL, network heterogeneity, long geographical distances, and the large size of encrypted data make the coordination a communication bottleneck. Thus, methods proposed to mitigate communication overhead typically involve reducing the cost of coordination and compressing the data transmitted between parties. We summarize these methods in Table \ref{tab:vfl_efficiency} and discuss them in this section.


\subsection{Multiple Client Updates} 

One straightforward way to save the communication cost is by allowing participating parties to perform multiple local updates during each iteration.  Liu et al.~\cite{liu2022fedbcd} proposed a federated stochastic block
coordinate descent algorithm, called FedBCD, that allows each party to conduct multiple client updates before each communication to reduce the number of synchronizations, thereby mitigating the communication overhead. 
Castiglia et al.~\cite{Castiglia2022FlexibleVF} proposed a flexible local update strategy for VFL, named Flex-VFL, that allows each party to conduct a different number of local updates constrained by a specified timeout for each communication round. 
Zhang et al.~\cite{Jie2022adavfl} proposed an adaptive local update strategy for VFL, named AdaVFL, that optimizes the number of local updates for each party in each round by minimizing the total training time. Xie et al.~\cite{xie2022improving} proposed an ADMM-based optimization method to implement multiple local updates. 
Fu et al.~\cite{fu2022communicationefficient} proposed CELU-VFL, an efficient VFL training framework that implements multiple local updates using cached statistics. {\blue{These methods typically require proper choices of training parameters, e.g.learning rate, to improve convergence and exhibit trade-off between computational resources and communication efficiency.}}

\subsection{Asynchronous Coordination} 
The core idea of asynchronous coordination is that each party can upload and download intermediate training results asynchronously. 
{\blue{However, asynchronous coordination may result in stale information, which may harm the overall model performance and jeopardize communication efficiency if the stale information is not dealt with properly}}.

Li et al.~\cite{li2020asynchronousVFL} proposed GP-AVFL that allows parties to update local models asynchronously by leveraging a gradient prediction technique to dynamically adjust local model gradients. Cai et al.~\cite{cai2022avfl} proposed AVFL that accelerates VFL training by omitting the updates from the slow parties with poor network conditions. 
Zhang et al.~\cite{zhang2022lowlatency} proposed a truncated VFL algorithm, called T-VFL, 
to discard parties with channel gains lower than a threshold.
Chen et al.~\cite{chen2020vafl} proposed a vertical asynchronous federated learning algorithm called VAFL, which utilizes a query-response strategy that decouples the coordination between the server and clients. Hu et al.~\cite{Hu2019FDMLAC} proposed FDML, allowing each party to update its local model asynchronously but based on the same sequence of randomly sampled training data.

$\text{AsySQN}$~\cite{zhang2021asysqn},
$\text{VFB}^2$~\cite{zhang2021secure}, and FDSKL~\cite{gu2020fdskl} all utilize a tree-structured communication scheme~\cite{zhang2018SVRG} to enhance the communication efficiency. $\text{AsySQN}$~\cite{zhang2021asysqn} additionally exploits approximated Hessian information to obtain a better descent direction. $\text{VFB}^2$~\cite{zhang2021secure} supports multiple active parties. 
FDSKL~\cite{gu2020fdskl} integrates a non-linear kernel method into vertical federated learning. It leverages the random features to approximate the kernel mapping function aiming to achieve efficient computation parallelism, and adopts doubly stochastic gradients to update the kernel function for scalability.
Han et al.~\cite{han2022fedgbf} employs the random forest (RF)~\cite{tin1995rdf} as the base learner for learning GBDT in order to enhance parallelism and save communication rounds. To reduce the long periods of idle time and accelerate the aggregation process under cryptography, VF$^{2}$Boost\cite{fu2021vf2boost} adopts a concurrent training protocol to take full advantage of computational resources and leverages a re-ordered accumulation technique and a histogram packing method to accelerate histogram construction and communication.

{\blue{
Asynchronous coordination may incur additional computation overhead for handling inconsistencies between the asynchronous updates. Thus, trade-offs between coordination and computation overhead should be carefully considered when applying asynchronous coordination methods.}}

\subsection{One-shot Communication} 

One-shot Communication alleviates communication overhead by coordinating only once during the entire training procedure. All proposed one-shot communication approaches follow a two-step training procedure: (1) All parties extract latent representations from their original data using unsupervised learning; (2) The active party trains the global model using these latent representations. 

Wu et al.~\cite{wu2022practical} proposed FedOnce, in which each party leverages an unsupervised learning method, called NAT (Noise As Targets)~\cite{Piotr2017nat}, to extract latent representations from its local data. Then the active party trains the global model using its local features combined with latent representations passed from passive parties. AE-VFL~\cite{cha2021implementing} leverages autoencoder to extract latent representations from each party's local data, while CE-VFL~\cite{khan2022communicationefficient} utilizes both Principal Component Analysis (PCA) and autoencoder to conduct the latent representation extraction. 

A trade-off for one-shot methods is that sample-wise representations of original data are permanently passed on to another party. Therefore, the privacy risks for revealing these representations need to be carefully evaluated, e.g., through inversion attacks or information theory studies. {\blue{Besides, one-shot methods typically involve computationally expensive unsupervised learning of effective representations. Therefore, the trade-off between communication and computation is worth investigating.}}
 
\subsection{Compression} 

Compression is a commonly used approach in VFL to alleviate communication overhead by reducing the amount of data transmitted among parties. {\blue{It can alleviate both communication and computation overheads, especially when expensive encryption operations (e.g., HE and MPC) are applied. }}

Neural network-based VFL algorithms naturally map high-dimensional input vectors to low-dimensional representations. Some works adopt specialized dimension-reduction techniques to compress data. AVFL~\cite{cai2022avfl} leverages Principle Component Analysis (PCA) to compress transmitted data, while CE-VFL~\cite{khan2022communicationefficient} utilizes both PCA and Autoencoders to learn latent representations from raw data. Two follow-up works, SecureBoost+~\cite{chen2021secureboost_plus} and eHE-SecureBoost~\cite{xu2021efficient}, of SecureBoost encode encrypted first-order and second-order gradients into a single message to {\blue{reduce the encryption operations and the size of data transmitted between parties, thereby saving communication bandwidth and computational costs}}. C-VFL \cite{castiglia2022compressed} allows an arbitrary compression scheme to be applied to VFL to enhance communication efficiency and provides theoretical analysis on the impact of compressor parameters. GP-AVFL~\cite{li2020asynchronousVFL} employs a double-end sparse compression (DESC) technique to save communication traffic volume by squeezing the sparsity in forward outputs of local models and backward gradients transmitted from the active party to passive parties. {\blue{Adaptive quantization techniques~\cite{yuzhuTIST2022,Jhunjhunwala2021,Linqi2022} may also be considered in future VFL research.}}

\subsection{Sample and Feature Selection}

{\blue{Another approach to improve communication efficiency is to reduce the amount of data used for training and inference. For example, Coreset-VFL~\cite{NEURIPS2022_coreset} constructs a coreset of samples to alleviate the communication burden, while FedSDG-FS~\cite{li2023fedsdg}, SFS-VFL~\cite{Zhang2022secfs}, LESS-VFL\cite{castiglia2023less}, FEAST~\cite{fu2023feast} and VFLFS~\cite{feng2022vflfs} filter out unimportant features to save communication costs.}} 



\section{Improving Effectiveness}
\label{effective}

Conventional VFL is only able to utilize aligned labeled samples. However, real-world applications often have limited \textit{aligned samples}, especially as the number of parties grows. The availability of labeled samples is also scarce in many cases, resulting in unsatisfactory performance. Moreover, the collaborative inference is required since each party only has a sub-model after training.


\begin{figure}[!ht]
\centering
\includegraphics[width=0.55\linewidth]{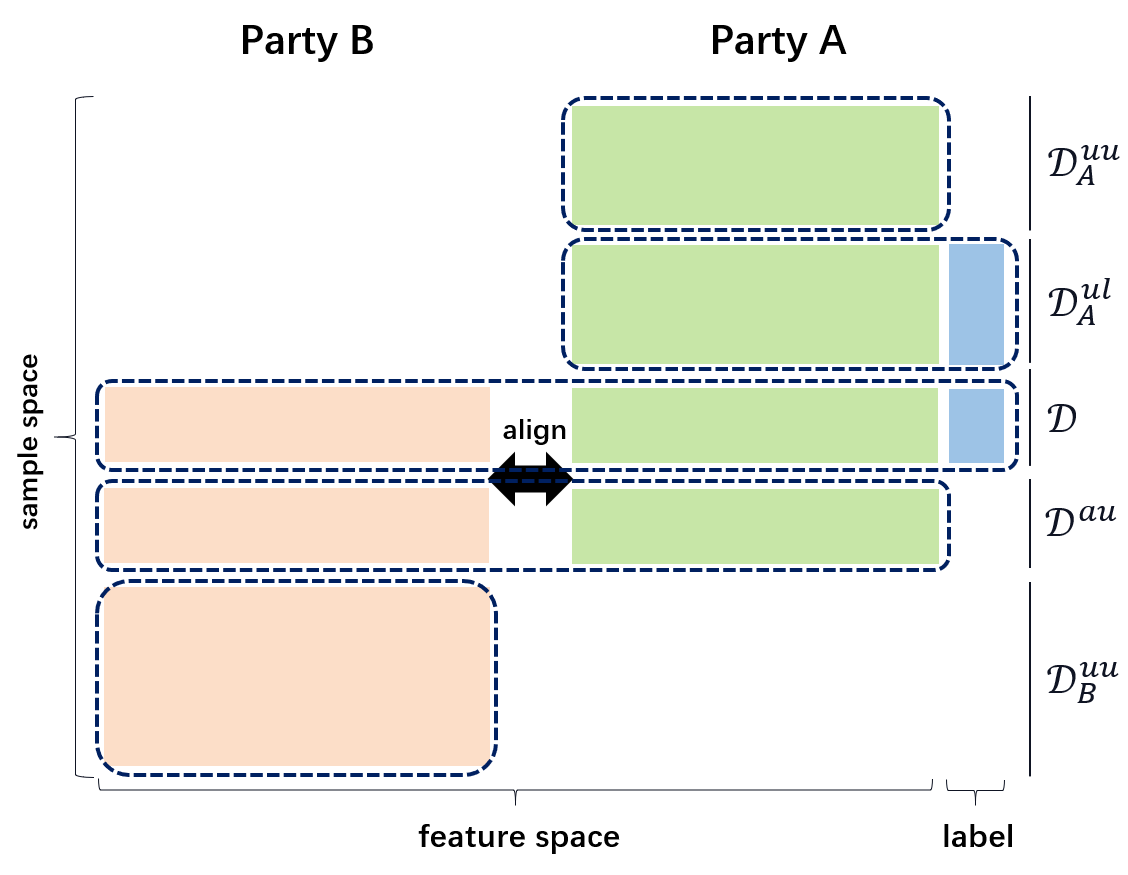}
\caption{The virtual dataset of a two-party VFL. \blue{$\mathcal{D}$ denotes the labeled and aligned samples used by the conventional VFL formulated in Eq. (\ref{eq:problem}), whereas $\mathcal{D}^{au}$ denotes aligned but unlabeled samples. $\mathcal{D}_A^{uu}$ and $\mathcal{D}_B^{uu}$ denote unaligned and unlabeled samples of party A and party B, respectively. $\mathcal{D}_A^{ul}$ denotes 
unaligned and labeled samples of party A.}}
\label{fig:VFL_dataset}
\end{figure}


To address these limitations, the literature has proposed various directions toward better utilizing available data to build a joint VFL model or helping participating parties build local predictors. 

For brevity, we discuss existing works through a two-party VFL setting involving an \textit{active party} $A$ and a \textit{passive party} $B$. We summarize these works in Table \ref{table:threat-models} and discuss them in the rest of this section. To better explain these works, we depict a general virtual dataset formed by the two parties (see Figure \ref{fig:VFL_dataset}). We dissect this virtual dataset into several sub-datasets to illustrate which portions of the virtual dataset are utilized by a VFL algorithm to train models, as reported in Table \ref{table:threat-models}. Specifically, $\mathcal{D}$ denotes the labeled and aligned samples, which is used by the conventional VFL formulated in Eq. (\ref{eq:problem}), whereas $\mathcal{D}^{au}$ denotes aligned but unlabeled samples. $\mathcal{D}_A^{uu}$ and $\mathcal{D}_B^{uu}$ denote unaligned and unlabeled samples of party A and party B, respectively. $\mathcal{D}_A^{ul}$ denotes 
unaligned and labeled samples of party A. 

\begin{table*}[!ht]
	\caption{Summary of existing works that aim to improve the effectiveness of VFL. Semi-SL, Self-SL, KD, and TL represent semi-supervised learning, self-supervised learning, knowledge distillation, and transfer learning, respectively. $\surd$ indicates its corresponding portion of data (see Figure \ref{fig:VFL_dataset}) is utilized by a specific VFL algorithm. \blue{Note that VFed-SSD has two objectives: one is to build a local predictor for the active party, and another is to build a joint predictor.}} 
	\centering
\scriptsize
	\begin{tabular}{c|c|c||c|c|c|c|c|c|c}
	    \hline
		\multirow{3}{*}{\shortstack{Core \\ Approach}} & \multirow{3}{*}{Objective} & \multirow{3}{*}{\shortstack{Existing \\ Work}} & \multicolumn{5}{c|}{Data Used} & \multirow{3}{*}{\shortstack{Method}} & \multirow{3}{*}{\shortstack{Party}} \\
		\cline{4-8}
		\\[-1em]
		~ & ~ & ~ & \multicolumn{2}{c|}{Aligned} & \multicolumn{3}{c|}{Unaligned} & ~ & ~ \\
			\\[-1em]
		\cline{4-5}	\cline{6-8}
		~ & ~ & ~ & $\mathcal{D}^{au}$ & $\mathcal{D}$ & $\mathcal{D}^{uu}_{B}$ & $\mathcal{D}^{uu}_{A}$ & $\mathcal{D}^{ul}_{A}$ & ~ & ~ \\
		\\[-1em]
	  \hline
        \hline
        \\[-1em]
	   \multirow{4}{*}{Self-SL} & \multirow{6}{*}{\shortstack{Build a joint \\ predictor}} & VFLFS~\cite{feng2022vflfs} & - & $\surd$ & $\surd$ & $\surd$ & - & \tiny Generative Models & $\geq 2$  \\
	   ~ & ~ & VFed-SSD~\cite{li2022semi} & $\surd$ & $\surd$ & - & - & - &  \tiny Contrastive Learning & $2$  \\
       ~ & ~ & FedHSSL~\cite{he2022hybrid} & $\surd$ & $\surd$ & $\surd$ & $\surd$ & - & \tiny Contrastive Learning & $\geq 2$  \\
       ~ & ~ &SS-VFNAS~\cite{liang2021selfsupervised} & - & $\surd$ & $\surd$ & $\surd$ & - & \tiny Contrastive Learning & $\geq 2$  \\
      \cline{1-1}
	   \multirow{2}{*}{Semi-SL} & ~ & FedCVT~\cite{kangyan2022fedcvt} & - & $\surd$ & $\surd$ & - & $\surd$ & \tiny Feature \& Label Estimation & $2$  \\
       ~ & ~ &FedMC~\cite{Yitao2022multiview} & - & $\surd$ & $\surd$ & - & $\surd$ &\tiny  Data Collaboration & $2$  \\
       \hline
	   \\[-1em]
       \multirow{4}{*}{KD} & \multirow{4}{*}{\shortstack{Build a local \\ predictor for \\ active party A}} & VFedTrans~\cite{wangvertical} & $\surd$ & - & - & - & $\surd$ & \tiny FedSVD \& Representation Distillation & $\geq 2$  \\
       ~ & ~ & VFL-Infer\cite{ren2022improving} & - & $\surd$ & - & - & - & \tiny Model Distillation & $2$  \\
	   ~ & ~ & VFed-SSD~\cite{li2022semi} & $\surd$ & $\surd$ & - & - & - & \tiny Model Distillation & $2$  \\
	   ~ & ~ & VFL-JPL~\cite{li2022vertical} & - & $\surd$ & - & - & $\surd$ & \scriptsize \tiny Feature Estimation \& Model Distillation & $2$  \\
	     \\[-1em]
		\hline
	     \\[-1em]
	     \multirow{4}{*}{TL} & \multirow{4}{*}{\shortstack{Build a local \\ predictor for \\passive party B}} & MMVFL~\cite{feng2020multi} & - & $\surd$ & - & - & - & \tiny Feature Selection \& Label Transfer & $\geq 2$  \\
		 ~ & ~& SFHTL~\cite{feng2022semisupervised} & - & $\surd$ & $\surd$ & - & $\surd$ & \tiny Feature \& Label Transfer & $\geq 2$  \\
		 ~ & ~& SFTL~\cite{YangLiu2019Secure,Sharma2019SecureAE} & $\surd$ & $\surd$ & - & - & $\surd$ & \tiny Feature Transfer & $2$  \\
		 ~ & ~& PrADA~\cite{kang2022prada} & $\surd$ & $\surd$ & - & - & - & \tiny Adversarial Domain Adaptation & $3$  \\
		 \hline
     \hline
	\end{tabular}
\label{table:threat-models}
\end{table*}

\subsection{Self-supervised Approaches}


Recently, self-supervised learning (Self-SL) has been introduced to VFL to improve the performance of the VFL model by exploiting unlabeled samples, which are not used in the conventional VFL. For illustrative purposes, we consider a two-party VFL scenario and rewrite Eq. (\ref{eq:problem}) as follows: 
\begin{align}\label{eq:ssl_vfl}
\hspace{-0.2cm} & \min_{\psi_A,\theta_A,\theta_B} \ell_{\text{VFL}}(\psi_A,\theta_A,\theta_B; \mathcal{D})
\end{align}

Self-SL-based VFL approaches proposed in the literature typically train participating parties' models $\psi_A$, $\theta_A$, and $\theta_B$ by minimizing a Self-SL loss based on unlabeled samples in addition to the main task loss defined in Eq. (\ref{eq:ssl_vfl}). We formulate a general Self-SL objective in VFL as follows:
\begin{align}\label{eq:self_sl}
\begin{split}
\tilde{\psi_A},\tilde{\theta}_A,\tilde{\theta}_B =
\argminF_{\psi_A,\theta_A,\theta_B} \ell_{\text{Self-SL}}(\psi_A,\theta_A,\theta_B;\mathcal{D}^{au},\mathcal{D}^{uu}_{A},\mathcal{D}^{uu}_{B})
\end{split}
\end{align}
where $\ell_{\text{Self-SL}}$ is the self-supervised learning loss that optimizes $\psi_A$, $\theta_A$ and $\theta_B$ for learning good representations using unlabeled data. 
Li et al.~\cite{li2022semi} proposed VFed-SSD that pretrains local models $\psi_A$, $\theta_A$ and $\theta_B$ through Eq. (\ref{eq:self_sl}) based on positive and negative sample pairs, which are formed from aligned data $\mathcal{D}^{au}$ leveraging matched pair detection (MPD) technique. Then, VFed-SSD finetunes pretrained models $\tilde{\psi}_A$, $\tilde{\theta}_A$ and $\tilde{\theta}_B$ through Eq. (\ref{eq:ssl_vfl}) based on labeled and aligned samples $\mathcal{D}$.
He et al.~\cite{he2022hybrid} proposed FedHSSL, a federated hybrid self-supervised learning framework, that pretrains $\theta_A$ and $\theta_B$ through Eq. (\ref{eq:self_sl}) based on cross-party views of aligned samples $\mathcal{D}^{au}$ and local views (via data augmentations) of unlabeled local samples $\mathcal{D}^{uu}_{A}$ and $\mathcal{D}^{uu}_{B}$. Then, FedHSSL finetunes $\psi_A$ and pretrained models $\tilde{\theta}_A$ and $\tilde{\theta}_B$ through Eq. (\ref{eq:ssl_vfl}) based on $\mathcal{D}$.
Feng~\cite{feng2022vflfs} proposed a VFLFS algorithm that optimizes Eq. (\ref{eq:ssl_vfl}) and Eq. (\ref{eq:self_sl}) in an end-to-end manner. It trains local models $\theta_A$ and $\theta_B$ using autoencoders based on unaligned data $\mathcal{D}^{uu}_{A}$ and $\mathcal{D}^{uu}_{B}$, and simultaneously finetunes these local models and the global module $\psi_A$ based on labeled aligned samples $\mathcal{D}$. 

\subsection{Semi-supervised Approaches}
Rather than boosting representation learning capability leveraging self-supervised learning, Kang et al.~\cite{kangyan2022fedcvt} and Yitao et al.~\cite{Yitao2022multiview} proposed semi-supervised learning approaches that augment labeled and aligned samples $\mathcal{D}$ to boost the performance of the VFL model. We formulate a general Semi-SL-based VFL objective as follows:
\begin{align}\label{eq:semi_sl}
\begin{split}
\min_{\psi_A,\theta_A,\theta_B,\tilde{\mathcal{D}}} & \ell_{\text{VFL}}(\psi_A,\theta_A,\theta_B; \tilde{\mathcal{D}})
+ \lambda \ell_{\text{Semi-SL}}(\psi_A,\theta_A,\theta_B;\mathcal{D},\mathcal{D}^{ul}_{A},\mathcal{D}^{uu}_{B})
\end{split}
\end{align}
where $\ell_{\text{Semi-SL}}$ is the semi-supervised learning loss that aims to expand $\mathcal{D}$ by pseudo-labeling unlabeled samples or adding newly labeled samples while achieving maximal stability and precision on labeling newly added samples. 

Kang et al.~\cite{kangyan2022fedcvt} proposed a Semi-SL algorithm named FedCVT to implement Eq. (\ref{eq:semi_sl}). More specifically, FedCVT estimates representations for missing features and predicts pseudo-labels for unlabeled samples to obtain an expanded training set, denoted as $\tilde{\mathcal{D}}$. To improve the quality of $\tilde{\mathcal{D}}$, FedCVT cherry-picks pseudo-labeled samples added to $\tilde{\mathcal{D}}$ through an ensemble approach. 
Then, FedCVT trains the VFL model based on $\tilde{\mathcal{D}}$. Yitao et al.~\cite{Yitao2022multiview} proposed FedMC that integrates data collaboration~\cite{imakura2020data} into VFL to implement Eq. (\ref{eq:semi_sl}). FedMC first forms a latent feature space using $\mathcal{D}$. In this latent feature space, it measures the distance between each pair of unaligned samples from the active party and passive party, respectively. Then, FedMC aligns two samples in a pair and adds aligned samples to $\mathcal{D}$ if their distance is less than a threshold to form expanded training set $\tilde{\mathcal{D}}$. Next, FedMC trains the VFL model based on $\tilde{\mathcal{D}}$.



\subsection{Knowledge Distillation-based Approaches}
In conventional VFL, the active party $A$ cannot make inferences alone, which limits the availability of the active party's prediction service. Some studies~\cite{wangvertical,ren2022improving,li2022semi,li2022vertical} proposed methods to help party $A$ build a local predictor instead of a VFL model while still benefiting from VFL training. To this end, they typically leverage Knowledge Distillation (KD) techniques to transfer knowledge of teacher models obtained through VFL to party $A$'s local models for enhancing performance. We formulate a general knowledge distillation-based VFL objective as follows.
\begin{align}\label{eq:ssl_dl}
\begin{split}
\min_{\mathbf \psi_A^s,\theta_A^s} & \ell_{A}(\psi_A^s,\theta_A^s; \mathcal{D}^{ul}_{A}) + \lambda \ell_\text{KD}(\psi_A^s,\theta_A^s,\psi_A^t,\theta_A^t, \theta_B^t;\mathcal{D}^{au})
\end{split}
\end{align}
where $\ell_{\text{KD}}$ is the knowledge distillation loss that forces to transfer knowledge from teacher models $\psi_A^t$, $\theta_A^t$ and $\theta_B^t$ to party $A$'s local models $\psi_A^s$ and $\theta_A^s$, $\ell_A$ is party A's task loss that optimizes $\psi_A^s$ and $\theta_A^s$ based on labeled samples $\mathcal{D}_A^{ul}$, and $\gamma$ is the hyperparameter that controls the strength of KD. $\psi_A^t$, $\theta_A^t$ and $\theta_B^t$ can be pretrained through Eq. (\ref{eq:ssl_vfl}) or Eq. (\ref{eq:self_sl}).
Wang et al.~\cite{wangvertical} proposed a vertical federated knowledge transfer approach (VFedTrans) via representation distillation that enables the active party $A$ to make inferences on unaligned local data. To this end, VFedTrans first learns federated representations through FedSVD~\cite{chai2022federated} based on aligned samples $\mathcal{D}^{au}$, and then it utilizes autoencoders as teacher models to transfer the knowledge encoded in the federated representations to the active party $A$'s local models $\psi_A^s$ and $\theta_A^s$ as students. Ren et al.~\cite{ren2022improving} proposed VFL-Infer, a VFL framework that pretrains teacher models $\psi_A^t$, $\theta_A^t$ and $\theta_B^t$ through Eq. (\ref{eq:ssl_vfl}), and then leverages these teacher models to help party $A$ train its local models $\psi_A^s$ and $\theta_A^s$ through Eq. (\ref{eq:ssl_dl}). 
Li et al.~\cite{li2022semi} proposed VFed-SSD that trains teacher models through Eq. (\ref{eq:self_sl}) using cross-party contrastive learning based on aligned data $\mathcal{D}^{au}$ and distills knowledge from teacher models to help the active party $A$ to train its local models $\psi_A^s$ and $\theta_A^s$. In another work along this line of research, Li et al.~\cite{li2022vertical} proposed a joint privileged learning in the VFL setting (VFL-JPL) to train local models for the active party $A$. By employing the feature imitation and ranking consistency restriction, VFL-JPL can effectively train the active party $A$'s local models through Eq. (\ref{eq:ssl_dl}) based on both aligned and unaligned samples as well as knowledge distilled from teacher models pretrained through Eq. (\ref{eq:ssl_vfl}).


\subsection{Transfer Learning-based Approaches}
Transfer-learning (TL) based VFL approaches~\cite{feng2020multi, YangLiu2019Secure, Sharma2019SecureAE, kang2022prada, feng2022semisupervised} treat the active party $A$ as the source domain with a large corpus of labeled samples and the passive party $B$ as the target domain with only unlabeled samples or a limited amount of labeled samples. These approaches leverage VFL as the bridge to transfer knowledge from party $A$ to party $B$. We formulate a general TL-based VFL objective as follows:
\begin{align}\label{eq:tl_vfl}
\begin{split}
\min_{\phi_B,\theta_B} & \ell_{B}(\phi_B;\theta_B;\mathcal{D}_B) + \lambda_1 \ell_{A}(\psi_A,\theta_A,\theta_B;\mathcal{D},\mathcal{D}^{ul}_{A}) + \lambda_2 \ell_{\text{TL}}(\theta_A,\theta_B;\mathcal{D}^{au},\mathcal{D}^{uu}_{A},\mathcal{D}^{uu}_{B})
\end{split}
\end{align}
where $\ell_{\text{TL}}$ is the transfer learning loss that aims to reduce the domain discrepancy between source and target domains, and $\ell_{A}$ is the source party $A$'s task loss that trains models using samples with labels of the source domain. $\ell_{\text{TL}}$ and $\ell_{A}$ together transfer the knowledge from the source domain to the target domain. The target party $B$ utilizes its task loss $\ell_{B}$ to further adapt the transferred knowledge to its local task using samples $\mathcal{D}_B$ with labels of the target domain if $\mathcal{D}_B$ is available. $\phi_B$ is the target party $B$'s local predictor. The target party $B$ may or may not need the help of party $A$ for inference, depending on the specific application of Eq. (\ref{eq:tl_vfl}).

Liu et al.~\cite{YangLiu2019Secure} proposed a secure federated transfer learning framework (SFTL), the pioneering work exploring transfer learning in VFL. SFTL first trains feature extractors $\theta_A$ and $\theta_B$ to map two heterogeneous feature spaces into a common latent subspace through aligned samples $\mathcal{D}^{au}$. In this latent subspace, the passive party $B$'s local models $\phi_B$ and $\theta_B$ are trained using data $\mathcal{D}_B$. 
As a follow-up work, Sharma et al.~\cite{Sharma2019SecureAE} leverage a more efficient secure computation framework named SPDZ~\cite{cryptoeprint:2011:535} to further enhance the efficiency of SFTL. 

SFTL can only transfer knowledge from one source party to one target party. To support multi-party knowledge transfer, Feng et al.~\cite{feng2020multi} proposed a Multi-Participant Multi-Class VFL (MMVFL) that leverages consistency regularization to transfer label information from the active party to all passive parties such that each passive party can learn a local predictor with its pseudo-labeled samples. Feng et al.~\cite{feng2022semisupervised} further proposed a semi-supervised federated heterogeneous transfer learning (SFHTL) that utilizes unaligned samples of all parties and aligned samples to build a local predictor for each party. Specifically, SFHTL utilizes an autoencoder to learn local representations from each party and then aggregates local representations to form global representations, through which labels of the active party are propagated to each passive party. With labeled local samples, each party can train its local predictor independently.

Kang et al.~\cite{kang2022prada} proposed PrADA to address the label deficiency of VFL through domain adaptation (DA). PrADA involves a label-rich source party $A$, a label-deficient target party $B$, and a third party that provides rich features for both parties $A$ and $B$. PrADA treats the third party as a bridge to transfer the knowledge from the source party to the target party and leverages the adversarial domain adaptation to minimize the domain discrepancy between the source and target domains.

\section{Preserving Data Privacy and Defending Against Attacks}

\label{privacy}
In a VFL system, privacy threats may emerge from the inside or the outside of the system, or both. 
If the attacker attempts to learn information about the private data of other parties without deviating from the VFL protocol, it is regarded as \textit{honest-but-curious}. The attacker is regarded as \textit{malicious} if it fails to adhere to the VFL protocol. 
In this section, we first review privacy-preserving protocols involved in the typical VFL framework (Sec. \ref{private entity alignment} and Sec. \ref{security_protocol}), followed by discussions on emerging research on attacks and defense strategies (Sec. \ref{sec:data leakage} and Sec. \ref{sec:backdoor}).  


\subsection{Private Entity Alignment} 
\label{private entity alignment}

\textbf{Private Set Intersection} (PSI) is the most common method for privacy-preserving entity alignment in VFL. In a PSI protocol, all parties cooperatively find the common ID intersection without revealing any information else. PSI protocols can be realized using various techniques, such as encryption and signature strategies \cite{liang2004privacy} and oblivious transfer \cite{Entity_OT, pinkas2018PSIOT} etc. The standard PSI protocol is typically applied to a two-party VFL system. \cite{zhou2021privacy,Lu2020MultiPSI} proposed methods for Entity Matching and PSI protocols that can be applied to multiple parties. PSI still reveals the common ID information. Several attempts have been made to enhance the privacy of the intersection ID set. \cite{liu2020asymmetrical} proposed an adapted PSI protocol for asymmetrical ID alignment using Pohlig-Hellman encryption scheme and a obfuscate set to help protect the entity information of a weaker party with far less samples than the other party from being exposed. 
\cite{sun2021vertical} proposed a method called FLORIST that safeguards the entity membership information for all parties by using a union ID set 
and generating synthetic data for missing IDs in the union set. However this method is limited to unbalanced binary classification tasks and incurs additional computational costs for generating and training the synthetic data. 


\subsection{Privacy-Preserving Training Protocols}
\label{security_protocol}
\begin{figure}[!ht]
\centering
\includegraphics[width=0.90\linewidth]{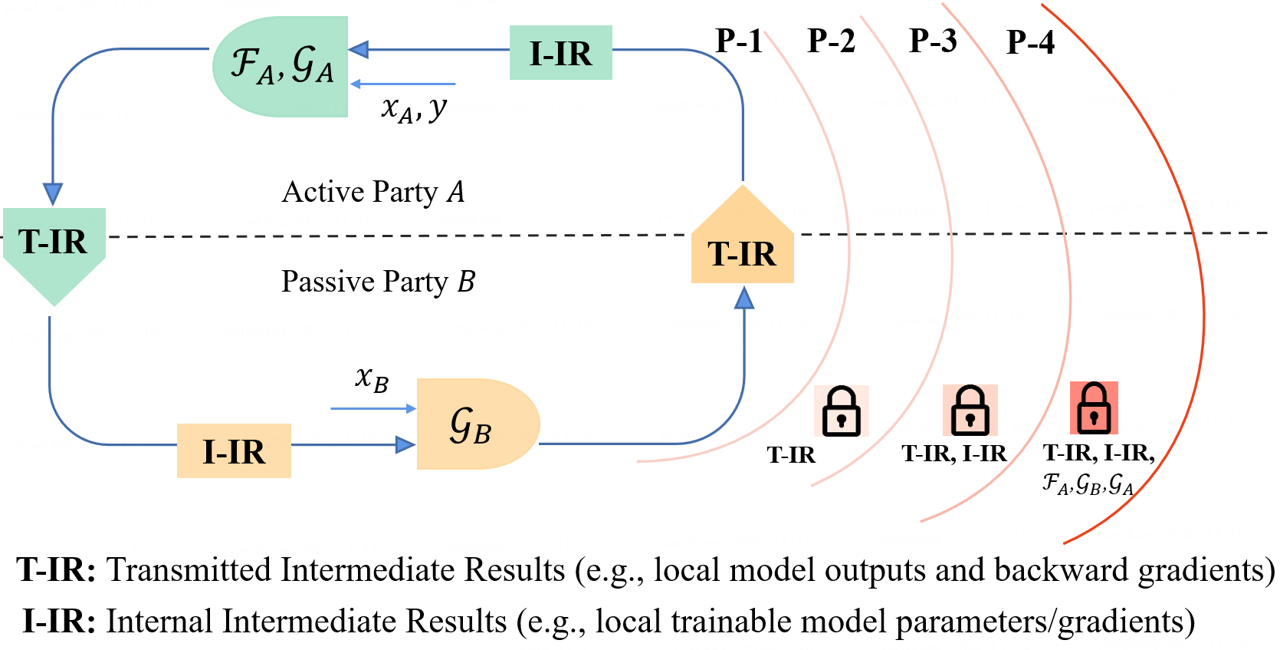}
\caption{A conceptual view on the information flowing within and between an active party $A$ and a passive party $B$ during training to illustrate security protocols $\text{P-1}$, $\text{P-2}$, $\text{P-3}$ and $\text{P-4}$.}
\label{fig:VFL_protocols}
\end{figure}

VFL approaches proposed in the literature adopt various security definitions and privacy-preserving protocols. In this section, we summarize these protocols based on what is protected and exposed during VFL training and inference. We first provide the basic protocol of VFL. We then discuss other protocols which adopt either relaxed or enhanced privacy constraints. Figure \ref{fig:VFL_protocols} illustrates these protocols.


\textbf{Basic Protocol (P-1): Keeping private data and models local.} All VFL participants keep their private data (e.g., labels and features), as well as the global module $\mathcal{F}_K$ and models $\{\mathcal{G}_k\}_{i=1}^K$ local during training and inference. Intermediate results are transmitted in plaintext for training and inference. We use this setting as our basic protocol (termed \textbf{P-1}). A case in point, during the training process of VFL (see Algo \ref{alg:first_order_VFL_he}), each party $k$'s intermediate results $H_k$ and gradients $\frac{\partial \ell}{\partial H_k}$ instead of raw data are transmitted, preventing private data from being revealed. Liu et al. \cite{liu2022fedbcd} provided security proof proving that private features $\mathbf{x}_k$ can not be exactly recovered in the P-1 protocol when no prior knowledge about data is available.  


\textbf{Relaxed Protocol (P-0): Nonprivate label or model.} In literature and applications, there are also cases where this security assumption of \textbf{P-1} is relaxed, resulting in a few variants of protocols, including:
\begin{itemize}
    \item Nonprivate Labels. These are cases where labels can be accessed by all parties for training and the security model is to protect features only \cite{Hu2019FDMLAC,castiglia2022compressed,Castiglia2022FlexibleVF}.
    \item Nonprivate global module or local models. These are cases where the global module~\cite{qiu2022lblrel} or local models~\cite{luo2021feature,he2019model,jiang2022comprehensive,jin2021cafe} are considered \textit{white-boxed} to adversaries.
\end{itemize}

Since these variants relax the basic security requirement of VFL, we assign a lower level to them (\textbf{P-0}), and we use \textbf{P-0(y)} and \textbf{P-0(g)} to denote the nonprivate label and nonprivate model scenarios, respectively. 
 
Building on the basic protocol \textbf{P-1}, privacy-preserving techniques have been adopted to further protect the training procedure, resulting in protocols with enhanced privacy. Below we describe the most representative protocols based on what is exposed, in ascending order of privacy level.  

\textbf{Standard Protocol (P-2): Protecting transmitted intermediate results}. In this protocol, P-1 is satisfied. In addition, the intermediate results transmitted between parties are protected by cryptography protocols, while other training information processed within each party is left in plaintext to balance privacy and efficiency. For example, HE~\cite{Yang2019FLconcept,Hardy2017PrivateFL} can be adopted to encrypt sample-level outputs $H_k$ and gradients $\frac{\partial \ell}{\partial H_k}$ transmitted between each passive party $k$ and the active party to thwart privacy attacks. Batch-level gradients $\nabla_{\theta_k} \ell$ computed within party $k$ are in plaintext for efficient training. The SecureBoost~\cite{SecureBoost} is another example where HE is used to protect transmitted intermediate results, but the aggregated gradients are exposed to the active party.


\textbf{Enhanced Protocol (P-3): Protecting entire training protocol}. In this protocol, P-2 is satisfied. In addition, no training information is revealed to any party except for the resulting trained models. For example, batch-level information such as local model gradients $\nabla_{\theta_k} \ell$ and parameters $\theta_k$ can be protected by adopting Secure Multi-Party Computation (MPC)~\cite{chen2021homomorphic}. Most existing works focus on the \textit{honest-but-curious} assumption, which assumes that the adversary follows the VFL protocol. To further handle malicious settings, more advanced privacy-preserving techniques such as SPDZ~\cite{Sharma2019SecureAE} have also been integrated with VFL \cite{Sharma2019SecureAE,sharma2019aaai}. 


\textbf{Strict Protocol (P-4): Protecting training protocol and learned models.} This protocol further enhances the P-3 to protect final learned models using privacy-preserving techniques such as secret sharing~\cite{pivot2020} and hybrid schemes that combine HE and SS~\cite{blindFL,FATELR}. It only reveals the final inference results but nothing else. This protocol addresses the emerging privacy challenge that the local model is exploited by its owner to infer private information about other parties \cite{fu2021label,SecureBoost,pivot2020}. However, it requires complex computations which limits its efficiency and scalability.

\begin{table*}[!ht]
	\caption{Summary of existing data inference attacks in VFL. A.P. represents the Attacking Phase. In the A.P. column, TRG denotes Training Phase and INF denotes Inference Phase.} 
	\centering
	\scriptsize
	\begin{tabular}{m{1.0cm}<{\centering}||m{4.5cm}<{\centering}|m{1.5cm}<{\centering}|m{1.2cm}<{\centering}|m{1.0cm}<{\centering}|m{0.7cm}<{\centering}|m{1.6cm}<{\centering}}
	    \hline
		~ & \multirow{1}*{\shortstack{Attacking Method}} & \shortstack{VFL \\ Setting} & \shortstack{Model} & \shortstack{Against \\ Protocol} & \shortstack{A.P.} & \shortstack{Auxiliary \\ Requirement} \\
	    \hline
	    \hline
		\multirow{13}*{\shortstack{Label \\Inference\\Attack}} &  Direct Label Inference (DLI)~\cite{fu2021label,zou2022defending} & aggVFL & NN & P-1 & TRG &  \textendash \\
        \cline{2-7}
		~ & \multirow{1}*{Norm Scoring (NS)~\cite{li2021label}} & $\text{splitVFL}_c$  & NN & P-1 & TRG &  \textendash \\
	    \cline{2-7}
		~ & \multirow{1}*{Direction Scoring (DS)~\cite{li2021label}} & $\text{splitVFL}_c$  & NN & P-1 & TRG &  \textendash \\
        \cline{2-7}
		~ & \multirow{1}*{Residual Reconstruction (RR)~\cite{tan2022residuebased}} & aggVFL & LR & P-2 & TRG  &  \textendash  \\
        \cline{2-7}
		~ & \multirow{1}*{Gradient Inversion (GI)~\cite{zou2022defending}} & aggVFL & NN & P-2 & TRG &  \textendash \\
	    \cline{2-7}
		~ & Gradient Inversion with a Label Prior ~\cite{sanjay2021exploit} & $\text{splitVFL}_c$ & NN & P-2 & TRG &  \scriptsize Label prior distribution \\
	   	\cline{2-7}
		~ & Passive Model Completion (PMC)~\cite{fu2021label} & splitVFL & NN & P-3 & INF  & \scriptsize Labeled data  \\
        \cline{2-7}
		~ & Active Model Completion (AMC)~\cite{fu2021label} & splitVFL & NN & P-3 & INF  & \scriptsize Labeled data  \\
        \cline{2-7}
		~ & \multirow{1}*{Spectral Attack (SA) ~\cite{sun2022label}} & $\text{splitVFL}_c$ & NN & P-3 & INF & - \\
        \cline{2-7}
		~ & Label-related Relation Inference (LRI) ~\cite{qiu2022lblrel} & $\text{splitVFL}_c$ & GNN & P-0(g) & INF &  - \\

        \hline
        \multirow{14}*{\shortstack{Feature \\Inference\\Attack}} & Binary Feature Inference (BFI)~\cite{peng2022binary} & splitVFL & NN & P-1 & TRG & \scriptsize Binary features \\
        \cline{2-7}
        ~ & Reverse Multiplication Attack  (RMA)~\cite{weng2020privacy} & aggVFL  & LR & P-2  & TRG  & \scriptsize Corrupted coordinator  \\
        \cline{2-7}
      \\ [-1em]
        ~ & Protocol-aware Active Attack (PAA)~\cite{hu2022vertical} & aggVFL & LR & P-2 & TRG  & \scriptsize Victim has 1 feature  \\
        \cline{2-7}
      \\ [-1em]
        ~ & \multirow{1}*{Reverse Sum Attack (RSA)~\cite{weng2020privacy}} & aggVFL & GBDT & P-2 & TRG  & \textendash  \\
        \cline{2-7}
    	~ & \multirow{1}*{Equality Solving Attack (ESA)~\cite{luo2021feature}} & aggVFL & LR & P-0(g) & INF & \textendash \\
        \cline{2-7}
		~ & \multirow{1}*{Path Restriction Attack (PRA)~\cite{luo2021feature}} & aggVFL & Tree & P-0(g) & INF & \textendash \\
        \cline{2-7}
		~ & Generative Regression Network (GRN)~\cite{luo2021feature} & aggVFL & NN & P-0(g) & INF & \textendash \\
        \cline{2-7}
		~ & White-Box Model Inversion (MI) ~\cite{he2019model,jiang2022comprehensive} & aggVFL / $\text{splitVFL}_c$ & LR/NN & P-0(g) & INF & \textendash\\
        \cline{2-7}
		~ & Black-Box Model Inversion (MI) ~\cite{he2019model,jiang2022comprehensive} & aggVFL / $\text{splitVFL}_c$ & LR/NN & P-1 & INF & \scriptsize Labeled data \\
        \cline{2-7}
		~ & Catastrophic Data Leakage in VFL (CAFE)~\cite{jin2021cafe} & $\text{aggVFL}_c$ & NN & P-0(g) & TRA & \textendash\\
		\cline{2-7}
      \\ [-1em]
        ~ & {\blue{Infer Attribute from Representation (IAR)~\cite{song2020overlearning}}} & $\text{aggVFL}_c$ \& $\text{splitVFL}_c$ & NN & P-0(g) & INF & Attribute data\\
		\hline
	\end{tabular}
\label{table:attacks}
\end{table*}


\subsection{Defending against Data Inference Attacks}
\label{sec:data leakage}
\begin{figure}[!tb]
 \centering
 \includegraphics[width=0.99\linewidth]{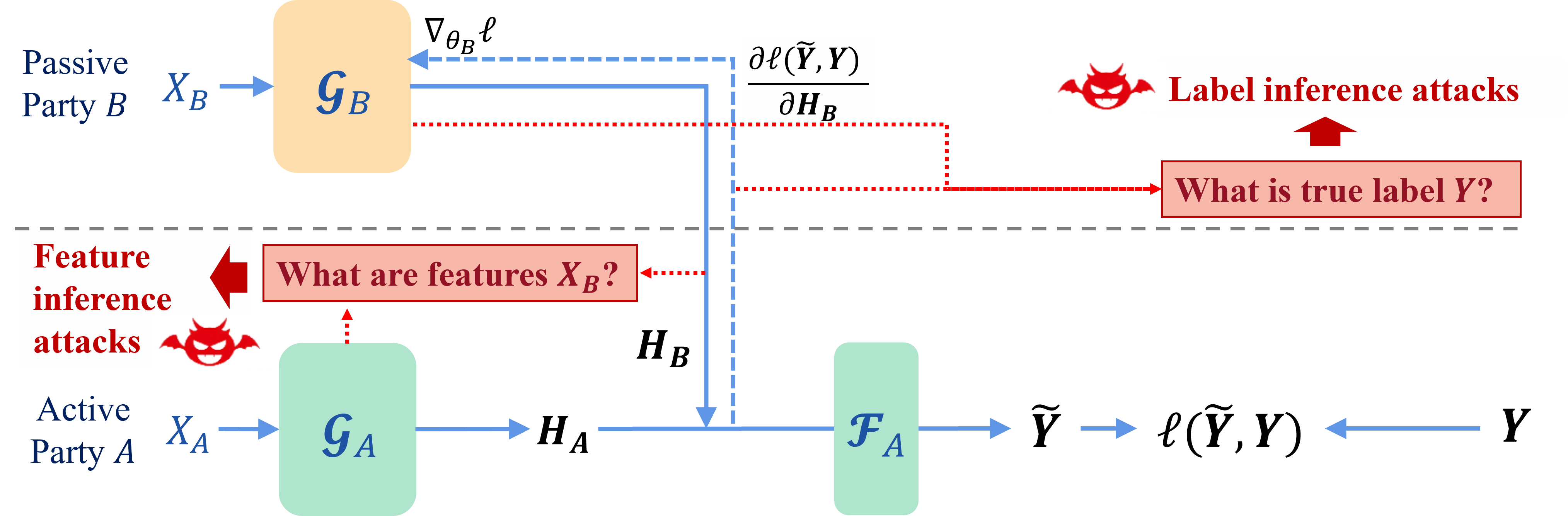}
 \caption{Illustration of data inference attacks in VFL system. The active party $A$ typically infers features or attributes of the passive party $B$, while the passive party $B$ typically infers labels of the active party $A$.} 
 \label{fig:sub_fig_vfl_attack_data_inference}
\end{figure}

In a typical VFL system, both features and labels are considered private, whereas most data attacks to HFL scenarios consider features as the target.  Therefore, both feature and label protections are critical research subjects for VFL. {\blue{Figure \ref{fig:sub_fig_vfl_attack_data_inference} illustrates data inference attacks in VFL.}}

\subsubsection{Label Inference Attacks}\label{sec:label_attack}

In real-world scenarios, labels such as patients' diagnostic results and individual loan default records are considered sensitive information that only authorized institutions can access.  A passive party $B$ (i.e., the attacker) may try to infer the valuable label owned by the active party $A$ using the information they accumulate during training or inference. It may follow the protocol \textit{passively} under the \textit{honest-but-curious} security assumptions or \textit{actively} by tampering with the protocol under the \textit{malicious} assumptions. The literature has proposed various label inference attacks under various security protocols, as summarized in Table \ref{table:attacks}. 

 


\textbf{Label inference attacks using sample-level gradient.} When the VFL applies P-1 protocol, a passive party $B$ (i.e., the attacker) has access to sample-level gradients $\frac{\partial \ell}{\partial H_B}$ sent backward from the active party $A$. The attacker can exploit this information to conduct Direct Label Inference (DLI) \cite{li2021label,fu2021label}. DLI can achieve accuracy up to $100\%$ if the active party adopts a \textit{nontrainable} global module $\mathcal{F}_A$ such as a softmax function because the gradient vector for each sample has only one element that has an opposite sign against all the others, thereby disclosing the labels~\cite{li2021label}. For special scenarios like binary classification, the attacker can deduce labels from sample-level gradients by mounting Norm Scoring (NS) or Direction Scoring (DS) attack~\cite{li2021label} even when the global module $\mathcal{F}_A$ is a trainable model (e.g., neural network). 

\textbf{Label inference attacks using batch-level gradients.} When the VFL applies the P-2 protocol, no intermediate result exchanged among parties is revealed to any party (e.g., encrypted by HE~\cite{Hardy2017PrivateFL}). Thus, the passive party $B$ (i.e., the attacker) cannot obtain sample-level gradients $\frac{\partial \ell}{\partial H_B}$, but it may have access to batch-level (i.e., local model) gradients $\nabla_{\theta_B} \ell$. Studies have shown that it is still possible to infer the true labels with high accuracy through the gradient inversion attack (GI)~\cite{zou2022defending, sanjay2021exploit} or the residue reconstruction attack (RR)~\cite{tan2022residuebased} using only the local model gradient. Following the same philosophy of the deep leakage from gradient method~\cite{zhu2019dlg}, passive party $B$ leverages GI to reconstruct the active party's labels by minimizing the distance between the predicted local model gradients $\nabla_{\theta_B} \hat{\ell}$ and the ground truth ones $\nabla_{\theta_B} \ell$. We formulate a general GI attack for inferring labels as follows:
\begin{align}\label{eq:gi_label_inf}
y^{*} = 
\argmin_{\hat{y}, \mathcal{\hat{\psi}}_A, \hat{H}_A} \ell_{\text{GI}}(\nabla_{\theta_B} \hat{\ell}, \nabla_{\theta_B} \ell) + \lambda R_{\text{GI}}(\hat{y})
\end{align}
where $\nabla_{\theta_B} \hat{\ell}= \nabla_{\theta_B}\mathcal{L}(\mathcal{F}_A(\hat{\psi}_A;\hat{H}_A,H_B),\hat{y})$, in which $\hat{y}$ is the label variable needs to be optimized, while $\mathcal{\hat{\psi}}_A$ and $\hat{H}_A$ are active party $A$'s global module parameter and local model output, respectively, that are estimated by party $B$ in order to mount GI attack because party B has no access to them; $\nabla_{\theta_B} \ell =  \nabla_{\theta_B}\mathcal{L}(\mathcal{F}_A(\psi_A;H_A,H_B),y)$ denotes the ground truth local model gradients; $R_{\text{GI}}$ regularizes the label variable $\hat{y}$ based on the label prior, aiming to enhance the quality of $\hat{y}$~\cite{sanjay2021exploit}.

The RR attack is tailored to linear models and aims to infer the plaintext value of encrypted gradient $[[\frac{\partial \ell}{\partial H_B}]]$ by solving an optimization problem as follows~\cite{kang2022eval}:
\begin{align}\label{eq:rr_attack}
\xi^* = \argminF_{\hat{\xi}} || x_B^{\text{T}} \cdot \hat{\xi} - \nabla_{\theta_B} \ell||^2_2
\end{align}
where $\hat{\xi}$ is the variable representing the plaintext value of $[[\frac{\partial \ell}{\partial H_B}]]$ and $\xi^*$ is the reconstructed values of $\frac{\partial \ell}{\partial H_B}$, based on which DLI, NS or DS can be applied to infer labels.

\textbf{Label inference attacks using trained models.} When the VFL applies the P-3 protocol, no training information is revealed to any party but only the final trained local model. The P-3 protocol can be achieved through MPC-based VFL approaches~\cite{chen2021homomorphic,FATELR}. A possible label inference strategy is for a passive party to finetune its trained local model with an inference head using auxiliary labeled data, and then predict labels using the complete model (i.e., the finetuned local model with the inference head). This attack is called Passive Model Completion (PMC)~\cite{fu2021label}, in which the passive party is semi-honest. An \textit{active} version of model completion (AMC) is also proposed in~\cite{fu2021label}. It leverages a malicious local optimizer instead of normal ones (e.g., Adam) to trick the trained federated model into relying more on the local model of the attacker than other parties such that the attacker can obtain a local model with better performance. MC relies heavily on the adequateness of the auxiliary data owned by the passive party as an attacker. Sun et al.~\cite{sun2022label} proposed a spectral attack (SA) that enables a passive party to predict labels by clustering outputs of the trained local model, thereby eliminating the dependency on auxiliary data. Qiu et al.~\cite{qiu2022lblrel} proposed a Label-related Relation Inference (LRI) attack targeting label-related relations in the graph owned by the active party, assuming the attacker has access to the global module and can obtain prediction results. LRI first recovers the active party's local outputs using an optimization-based method. It then recovers relations by forming an adjacency matrix based on outputs from the attacker's and the active party's local models and prediction results.


\subsubsection{Feature Inference Attacks}\label{sec:feature_attack}

An individual's original feature is at the heart of privacy protection because it contains sensitive information that is not allowed to share. Various attacking methods has been proposed to infer features from shallow models (e.g., logistic regression and decision trees)~\cite{weng2020privacy,luo2021feature, peng2022binary} and complex models (e.g., neural networks and random forests) ~\cite{luo2021feature,he2019model,jiang2022comprehensive, jin2021cafe}. We summarize existing feature inference attacks in Table \ref{table:attacks}. These attacks are typically under the setting where the active party (with labels) $A$ is the attacker who attempts to recover features of a passive party $B$. The attackers in proposed feature inference algorithms may or may not have the knowledge of the passive party's model parameters $\theta_B$, which are, respectively, referred to as the \textit{white-box} and \textit{black-box} settings.


\textbf{Feature inference attacks under white-box setting.} Under the white-box setting, the attacker (i.e., the active party or the server) has access to its own model $\mathcal{G}_A$, the passive party's local model $\mathcal{G}_B$, the aligned data indices and possibly labels. In literature, there are mainly two ways to conduct white-box feature inference attacks: model inversion~\cite{he2019model,jiang2022comprehensive} during the inference phase and gradient inversion during the training phase~\cite{jin2021cafe}. 

The core idea of model inversion (MI) is to optimize variable $\hat{x}_B$ to approximate the passive party's real input data $x_B$ such that the predicted output $\hat{v}$ of the VLF model is close enough to the real output $v$ computed based on $x_B$. We formulate a general MI attack as follows:
\begin{align}\label{eq:wb_feat_inf}
x_B^{*} = 
\argmin_{\hat{x}_B} \ell_{\text{MI}}(\hat{v}, v) + \alpha R_{\text{MI}}(\hat{x}_B)
\end{align}
where $\mathcal{L}_{\text{MI}}$ is the loss function that minimizes the distance between $v$ and $\hat{v}$ to optimize $\hat{x}_B$, and $R_{\text{MI}}$ regularizes the variable $\hat{x}_B$ based on a prior knowledge. $\hat{v}$ is computed by:
\begin{align}\label{eq:wb_feat_inf_2}
\hat{v} = 
\mathcal{F}_A\left(\mathcal{G}_B(\hat{x}_{B},\theta_B),\mathcal{G}_A(x_A,\theta_A), y_A\right)
\end{align}
where $x_A$ and $y_A$ are features and labels belonging to the active party; local models $\mathcal{G}_A$ and $\mathcal{G}_B$ can be linear models, tree models or neural network models, and their model parameters $\theta_A$ and $\theta_B$ are fixed during the optimization; $\mathcal{F}_A$ is the global module that aggregates the outputs of local models and generates $\hat{v}$.

He et al.~\cite{he2019model} and Jiang et al.~\cite{jiang2022comprehensive} proposed similar white-box model inversion attacks under the SplitNN and aggVFL settings, respectively. Luo et al.\cite{luo2021feature} proposed three white-box feature inference attacks to learn $\hat{x}_B$ for three different models. These attacks can be seen as specialized MI. More specifically, they designed an Equality Solving Attack (ESA) for the logistic regression, a Path Restriction Attack (PRA) for the decision tree, and a Generative Regression Network (GRN) for attacking the neural network and random forest. The three attacks generally follow the optimization problem defined in Eq. (\ref{eq:wb_feat_inf}).

The gradient inversion (GI) attack was initially proposed in ~\cite{zhu2019dlg} under the HFL setting. The CAFE~\cite{jin2021cafe} extended GI to a white-box VFL setting, where the attacker has access to the passive party's model parameters and gradients as well as the aligned data indices. With this knowledge, CAFE can achieve state-of-the-art data recovery quality even with large batch sizes.  



\begin{table*}[ht!]
\centering
\scriptsize
\caption{Summary of existing cryptographic defense strategies in VFL. In the Defense Scheme column, GC denotes Garbled Circuits, SS denotes Secret Sharing, HE denotes Homomorphic Encryption, FE denotes Functional Encryption and TEE denotes Trusted Execution Environment. In the Adversarial Assumption column, SH denotes Semi-Honest and MA denotes Malicious. In the Protocol column, we assign each defense with Protocols (see Sec. \ref{security_protocol}) it satisfies; "$a$" and "$p$" denote active and passive parties, respectively. } 
\begin{tabular}{m{2.3cm}<{\centering}||m{1.2cm}<{\centering}|m{1.1cm}<{\centering}|m{1.3cm}<{\centering}|m{1.8cm}<{\centering}|m{0.7cm}<{\centering}|m{1.6cm}<{\centering}|m{2.1cm}<{\centering}}
\hline
\multicolumn{1}{c||}{\shortstack{Defense \\ Work}}  &
\multicolumn{1}{c|}{\shortstack{VFL \\ Setting}}  &
\multicolumn{1}{c|}{\shortstack{Model}}   &
\multicolumn{1}{c|}{\shortstack{Defense \\ Scheme}}   &
\multicolumn{1}{c|}{\shortstack{Protocol}} & 
\multicolumn{1}{c|}{Party} & 
\multicolumn{1}{c|}{\shortstack{Require \\ Coordinator}} &
\multicolumn{1}{c}{\shortstack{Adversarial \\ Assumption}} \\

\hline
\hline
GasconLR~\cite{Gascon2016SecureLR} & aggVFL & LR & GC+SS & P-3 & $\geq 2$ & \cmark & SH \\
\hline
HardyLR~\cite{Hardy2017PrivateFL} & aggVFL & LR & HE & P-2 & $\geq 2$ & \cmark  & SH\\
\hline
BaiduLR~\cite{yang2019parallel} & aggVFL & LR & HE & P-2 & $\geq 2$ & \xmark & SH \\
\hline
SecureLR~\cite{he2022securelr} & aggVFL & LR & HE+SS & P-2 & $\geq 2$  & \xmark & SH\\
\hline
\multirow{1}*{CAESAR~\cite{chen2021homomorphic}} & aggVFL & \multirow{1}*{LR} & \multirow{1}*{HE+SS} & P-3 & \multirow{1}*{$=2$} & \xmark & SH \\
\hline
 \multirow{1}*{HeteroLR~\cite{FATELR}} & aggVFL & \multirow{1}*{LR} & \multirow{1}*{HE+SS} & $a:$P-3, $p:$P-4 & \multirow{1}*{$= 2$} & \xmark & SH \\
\hline
 \multirow{1}*{FedV~\cite{xu2021fedv}} & aggVFL & \multirow{1}*{LR/SVM} & \multirow{1}*{FE} & P-2 & \multirow{1}*{$\geq 2$} & \cmark & SH\\
\hline
 SecureBoost~\cite{SecureBoost} & aggVFL & XGB & HE & P-2 & $\geq 2$ & \xmark & SH\\
\hline
 SecureBoost+~\cite{chen2021secureboost_plus} & aggVFL & XGB & HE & P-2 & $\geq 2$ & \xmark & SH\\
\hline
 SecureXGB~\cite{wen2021securexgb} & aggVFL & XGB & HE+SS & P-3 & $= 2$ & \xmark & SH\\
\hline
 MP-FedXGB~\cite{xie2022mpfedxgb} & aggVFL & XGB & SS & P-3 & $\geq 2$ & \cmark & SH\\
\hline
 SecureGBM~\cite{feng2019securegbm} & aggVFL & LGBM & HE & P-2 & $= 2$ & \xmark & SH\\
\hline
Pivot~\cite{pivot2020} & aggVFL & RF / GBDT & HE+SS & P-3 & $\geq 2$ & \xmark & SH, $\leq K$-$1$ colluded parties\\
\hline
Enhanced Pivot~\cite{pivot2020} & aggVFL & DT & HE+SS & P-4 & $\geq 2$ & \xmark & SH, $\leq K$-$1$ colluded parties\\
\hline
FedSGC~\cite{cheung2021Fedsgc} & $\text{aggVFL}_c$ & GNN & HE & P-2 & $= 2$ & \xmark & SH\\
\hline
ACML~\cite{Zhang2020acml} & $\text{splitVFL}_c$ & NN & HE & P-1 & $= 2$ & \xmark & SH\\
\hline
PrADA~\cite{kang2022prada} & $\text{splitVFL}$ & NN & HE & P-1 & $\geq 2$ & \xmark & SH\\
\hline
\multirow{1}*{BlindFL~\cite{blindFL}} & $\text{splitVFL}$ & \multirow{1}*{NN} & \multirow{1}*{HE+SS} & $a:$P-2, $p:$P-4  & \multirow{1}*{$=2$}  & \xmark & SH\\
\hline
 SFTL~\cite{YangLiu2019Secure} & $\text{aggVFL}$ & NN & HE & P-2 & $= 2$ & \xmark & SH \\
 \hline
 SFTL~\cite{YangLiu2019Secure} & $\text{aggVFL}$ & NN & SS & P-3 & $= 2$ & \xmark & SH\\
\hline
 SEFTL~\cite{Sharma2019SecureAE} & $\text{aggVFL}$ & NN & HE+SPDZ & P-3  & $= 2$ & \xmark & MA,dishonest majority\\
\hline
 N-TEE~\cite{chamani2020mitigating} & $\text{aggVFL}$ & XGB & TEE & P-3  & $ \geq 2$ & \xmark & SH\\
\hline
\end{tabular}
\label{tab:crypto_defenses_label}
\end{table*}

\textbf{Feature inference attacks under black-box setting.} Attackers under the black-box setting typically have some prior knowledge about the model or data of the passive party in order to conduct feature inference successfully.

Peng et al.~\cite{peng2022binary} proposed a Binary Feature Inference attack (BFI) to reconstruct binary features from the passive party's local model output $H_B$ (P-1 protocol), assuming the local model only has one fully-connected layer. In addition, BFIA adopts the Leverage Score Sampling technique~\cite{michael2011randalgo} to boost the attack efficiency. Weng et al.~\cite{weng2020privacy} and Hu et al.~\cite{hu2022vertical} proposed a Reverse Multiplication Attack (RMA) and a Protocol-aware Active Attack (PAA), respectively, to infer the private features $x_B$ of the passive party $B$ in the vertical logistic regression setting that applies P-2 protocol. In RMA, the attacker infers features $x_B$ of the passive party $B$ by solving linear equations in which $x_B$ is the only unknown variable, assuming the coordinator helps decrypt ciphertexts. In PAA, the attacker first obtains the passive party $B$'s outputs through solving a linear system and then utilizes these outputs to infer features of the passive party $B$. Weng et al.~\cite{weng2020privacy} also proposed a Reserve Sum Attack (RSA) targeting SecureBoost. RSA aims to infer the partial order of the passive party's input features by encoding magic numbers into the least significant bits of the encrypted first and second-order gradients. He et al.~\cite{he2019model} proposed a black-box model inversion (MI) attack to learn $x_B^{*}$ under the splitNN setting. More specifically, the attacker first trains a shadow model $\hat{\mathcal{G}}_B$ that mimics the behavior of the local model $\mathcal{G}_B$ using some auxiliary data, and then the attacker learns $x_B^{*}$ according to Eq. (\ref{eq:wb_feat_inf}) and Eq. (\ref{eq:wb_feat_inf_2}) with $\hat{\mathcal{G}}_B$ in place of $\mathcal{G}_B$. Jiang et al.~\cite{jiang2022comprehensive} proposed a similar MI method under the aggVFL setting.

\textbf{Attribute Inference Attacks.} {\blue{Aside from original features, privacy-sensitive attributes not represented in training data may also be inferred through overlearned model~\cite{song2020overlearning}.}}

In the rest of this subsection, we discuss defense strategies that alleviate the threat posed by these attacks.

\subsubsection{Cryptographic Defense Strategies}\label{cds}

Cryptographic Defense Strategies (CDS) use secure computations to evaluate functions on multiple parties in a way that only the necessary information is exposed to intended participants while preventing private data from being inferred by possible adversaries. Today, large-scale deployment of CDS to machine learning models, especially deep learning models, is still challenging. The focus of existing works in this direction is to improve the privacy-efficiency trade-off through the in-depth designing of privacy-preserving protocols. We adopt protocols defined in Sec. \ref{security_protocol} as a vehicle to compare representative CDS, as listed in Table \ref{tab:crypto_defenses_label}. We consider a defense follow a particular protocol only when it satisfies all requirements of that protocol.

A line of research works~\cite{Hardy2017PrivateFL,Gascon2016SecureLR,yang2019parallel,chen2021homomorphic,Yang2019FLconcept,he2022securelr,FATELR} focuses on designing CDS to protect the data privacy of vertical linear and logistic regressions. Gascon et al.~\cite{Gascon2016SecureLR} proposed a hybrid MPC protocol that combines Yao's garbled circuits with tailored protocols for securely solving vertical linear regression (GasconLR). Hardy et al.~\cite{Hardy2017PrivateFL} proposed a HE-based scheme for training the vertical logistic regression (HardyLR). Follow-up works BaiduLR~\cite{yang2019parallel} and SecureLR~\cite{he2022securelr} remove the coordinator from the training and inference procedure by relaxing either efficiency or privacy constraint. HardyLR, BaiduLR and SecureLR are vulnerable to privacy attacks targeting batch-level gradients (Sec.~\ref{sec:label_attack}). To address this limitation, Chen et al.~\cite{chen2021homomorphic} proposed a hybrid defense, named CAESAR, that combines HE and MPC to encrypt all intermediate results during the training and inference phases except the resulting trained models. The HeteroLR module of FATE~\cite{FATELR} extends CAESAR further to encrypt the passive party's local model after training.

Designing CDS for vertical neural networks (VNN) is more challenging for both computation and communication. Therefore, current CDS for VNN either target shallow neural networks~\cite{YangLiu2019Secure,Sharma2019SecureAE,blindFL} or are tailored to protect specific intermediate results exposed to the adversary~\cite{kang2022prada,Zhang2020acml} for balancing privacy and efficiency. SFTL~\cite{YangLiu2019Secure} designed a HE-based protocol and an SS-based protocol, respectively, to encrypt information shared between two parties that adopt neural networks with one or two layers. The follow-up work~\cite{Sharma2019SecureAE} leverages SPDZ~\cite{cryptoeprint:2011:535} to enhance the efficiency of SFTL further. BlindFL~\cite{blindFL} is proposed to build privacy-preserving VNN models through a federated source layer (FSL), which leverages a hybrid scheme mixing HE and MPC to guarantee the privacy of original data. 
ACML~\cite{Zhang2020acml} is proposed to build privacy-preserving SplitVFL and introduces a HE-equipped interactive layer between the active party and the passive party to protect the passive party's local model output. PrADA~\cite{kang2022prada} extends the interactive layer of ACML to the splitVFL setting, in which the global module is a linear model and local models are neural networks. FedSGC~\cite{cheung2021Fedsgc} utilizes HE to protect transmitted graph structural information. 


For tree-based VFL, SecureBoost~\cite{SecureBoost}, SecureBoost+~\cite{chen2021secureboost_plus}, SecureXGB~\cite{wen2021securexgb}, and MP-FedXGB~\cite{xie2022mpfedxgb} integrate XGBoost into VFL. SecureBoost and SecureBoost+ exploit additive homomorphic encryption (HE) to encrypt the information transmitted between parties to protect private data. SecureXGB protects all intermediate results through a hybrid scheme combining additive HE and secret sharing (SS), thereby enhancing the privacy level. MP-FedXGB proposed a SS scheme with distributed optimization to support more-than-two-party scenarios. SecureGBM~\cite{feng2019securegbm} 
is a LightGBM-based VFL using additive HE to protect transmitted information. Pivot~\cite{pivot2020} utilizes SS mixed with additive HE to guarantee that no intermediate information is disclosed. It additionally proposed an enhanced protocol to conceal the values of leaf labels and split thresholds from all participating parties, as well as protocols to handle malicious parties. Targeting SecureBoost, Chamani et al.~\cite{chamani2020mitigating} introduced a feature inference attack leveraging approximate distribution of feature values and proposed two countermeasures based on Trusted Execution Environment (TEE) to mitigate feature leakage risks.  

CDS are typically applied to utility-critical applications, such as finance and healthcare, to achieve lossless model utility (i.e., performance) while maintaining an acceptable balance between privacy and efficiency. For applications in which efficiency is a major concern or CDS are not feasible, non-cryptographic defense strategies are preferred.

\begin{table*}[ht!]
\centering
\scriptsize
\caption{Summary of emerging specialized defense strategies for defending against data leakage attacks (see Table \ref{table:attacks}).}
\begin{tabular}{m{2.1cm}<{\centering}|m{2.0cm}<{\centering}||m{1.3cm}<{\centering}|m{0.8cm}<{\centering}|m{3.2cm}<{\centering}|m{1.2cm}<{\centering}|m{1.8cm}<{\centering}}
\hline
 \multicolumn{1}{c|}{}  &
 \multicolumn{1}{c||}{\shortstack{Defense \\ Work}}  &
\multicolumn{1}{c|}{\shortstack{VFL \\ Setting}}  &
\multicolumn{1}{c|}{\shortstack{Model}}   &
\multicolumn{1}{c|}{\shortstack{Defense Scheme}}   &
\multicolumn{1}{c|}{\shortstack{Against\\Attack}} & 
\multicolumn{1}{c}{\shortstack{Defending\\ Party}} \\

\hline
\hline
 \multirow{8}*{\shortstack{Defenses against \\ Label Inference \\ Attack}} & MARVELL\cite{li2021label} & $\text{splitVFL}_c$ & NN & Add Noise & NS, DS & Active party\\
\cline{2-7}
 ~ & Max-Norm\cite{li2021label} & $\text{splitVFL}_c$ & NN & Add Noise &  NS, DS  & Active party\\
\cline{2-7}
 ~ & CAE~\cite{zou2022defending} & aggVFL & NN & HE+Disguise Label & DLI, MC & Active party\\
\cline{2-7}
 ~ & DCAE~\cite{zou2022defending} & aggVFL & NN & HE+Disguise Label+DG & DLI, MC &  Active party\\
\cline{2-7}
 ~ & PELoss~\cite{zheng2022peloss} & $\text{splitVFL}_c$ & NN & Potential Energy Loss & MC & Active party\\
\cline{2-7}
 ~ & dCorr~\cite{sun2022label} & $\text{splitVFL}_c$ & NN & Minimize Correlation & SA &  Active party \\
\cline{2-7}
 ~ & RM~\cite{tan2022residue} & aggVFL & LR & HE+Random Mask & RR &  Active party \\
\hline
 \multirow{7}*{\shortstack{Defenses against \\ Feature Inference \\ Attack}} & FG~\cite{jin2021cafe} & splitVFL & NN & Random Fake Gradients & CAFE & Passive party\\
\cline{2-7}
 ~ & DRAVL~\cite{sun2021defending} & $\text{splitVFL}_c$ & NN & Adversarial Training & MI & Passive party\\
 \cline{2-7}
 ~ & MD~\cite{peng2022binary} & $\text{splitVFL}$ & NN & Masquerade & BFIA & Passive party\\
  \cline{2-7}
 ~ & DP-Paillier-MGD~\cite{hu2022vertical} & $\text{aggVFL}$ & LR & HE+DP & PAA & Passive party\\
 
 \cline{2-7}
\\[-1em]
 ~ & FedPass~\cite{Gu2023fedpass} & splitVFL & NN & Passport & CAFE, MI &  Passive party \\

\hline
\end{tabular}
\label{tab:emerging_defenses_lable}
\end{table*}

\subsubsection{Non-cryptographic Defense Strategies}

Non-cryptographic Defense Strategies preserve privacy essentially by reducing the dependence between private data and the information exposed to the attacker. There are several representative ways to reduce such dependency, including adding noise, gradient discretization \cite{dryden2016communication}, gradient sparsification \cite{aji2017sparse,lin2017deep} and their hybrid \cite{shokri2015privacy}.
These methods typically exhibit a trade-off between utility and privacy. 

\textbf{Adding Noise} (DP)\cite{zhu2019dlg, fu2021label,li2021label,dwork2008differential} is a basic defense method for reducing leakage in FL. Noise following Laplace distribution or Gaussian distribution is commonly used. In VFL settings, it typically adds noise to the gradients or intermediate results shared with other parties to defend against feature or label leakages \cite{fu2021label,liang2021selfsupervised}. \cite{wang2020hybrid} introduced a hybrid differentially private VFL method that adds Gaussian noise to all parties' intermediate results to achieve both local and joint differential privacy. ~\cite{tian2020federboost, li2022opboost} applied differentially private noise to federated gradient-based decision trees in customized ways to achieve a good privacy-utility trade-off. Chen et al.~\cite{chen2022verticalgraph} integrate GNN into splitVFL setting and leverage DP-enhanced additive secret sharing to protect data privacy. \textbf{Gradient Discretization} (GD)\cite{dryden2016communication} encodes originally continuous gradients into discrete ones, aiming to reduce the private information disclosed to the attacker so that the attacker cannot precisely infer private data through discrete gradients.~\cite{zou2022defending,fu2021label} leveraged a specialized version of GD, named DiscreteSGD, to defend against label inference attacks in VFL. \textbf{Gradient Sparsification} (GS)\cite{aji2017sparse} removes a portion of the original gradients with small absolute values by setting them to $0$ while preserving the convergence of the original VFL task. Similar to GD, GS leverages information reduction to mitigate privacy leakage. GS are readily applied to distributed learning and HFL scenarios~\cite{aji2017sparse,lin2017deep}. 
It is also effective in defending against various label inference attacks for VFL.~\cite{zou2022defending,fu2021label}. 



A feasible direction to achieve better trade-offs between privacy and utility is designing hybrid defense schemes combining multiple defense strategies~\cite{shokri2015privacy}. Another direction is to design specialized defense strategies tailored to specific data inference attacks.

\subsubsection{Emerging Specialized Defense Strategies}
\label{defense_emerge}

Emerging specialized defense strategies are designed to thwart attacks that are difficult to defend against by traditional defense strategies.
We compare representative emerging defense strategies in Table \ref{tab:emerging_defenses_lable}.

\textbf{Defenses against label inference attacks.} Li et al. proposed MARVELL~\cite{li2021label}, which is tailored to defend against Norm Scoring (NS) and Direction Scoring (DS) attacks by adding optimized noise to the sample-level gradients. They also proposed a heuristic Max-Norm defense against the two attacks. 
Liu et al.~\cite{zou2022defending} proposed label disguising methods, called Confusional AutoEncoder (CAE) and DiscreteSGD-enhanced Confusional AutoEncoder (DCAE), which directly protects label information by encoding the original real label to soft fake labels with maximum confusion. 
PEloss~\cite{zheng2022peloss} and dCorr~\cite{sun2022label} are two auxiliary losses that are proposed to defend against the Model Completion (MC) attack and Spectral Attack (SA), respectively. Both methods try to train the attacker's local model for a large generalization error. 
Tan et al.~\cite{tan2022residue} proposed a Random Masking (RM) defense against the Residue Reconstruction attack (RR) by injecting zeros into randomly selected positions of the HE-encrypted sample-level gradients to prevent the RR from reconstructing these gradients correctly. {\blue{FedPass~\cite{Gu2023fedpass} leverages passport techniques to thwart both label and feature inference attacks.}}

\textbf{Defenses against feature inference attacks.} Fake Gradients (FG)\cite{jin2021cafe} is proposed to defend against Catastrophic Data Leakage in VFL (CAFE) by replacing the true gradients with randomly generated ones while keeping their corresponding positions. Sun et al.~\cite{sun2021defending} proposed DRAVL to defend against Model Inversion (MI) through adversarial training. In~\cite{peng2022binary}, a Masquerade Defense (MD) is proposed to thwart the Binary Feature Inference attack (BFI) by misleading the attacker to focus on randomly generated binary features, thereby protecting the true binary features. Hu et al. proposed DP-Paillier-MGD~\cite{hu2022vertical} to thwart the Protocol-aware Active Attack (PAA) by masking encrypted sensitive information to prevent the attacker from learning the precise value of the passive party's output and, thereby, the private features. {\blue{Adversarial training ~\cite{jia2018attriguard,song2020overlearning} and mutual information regularization ~\cite{song2020overlearning} were proposed to safeguard sensitive attributes of training samples.}}


\subsection{Defending against Backdoor Attacks}
\label{sec:backdoor}
Different from data leakage attacks, whose target is to invade privacy and steal data, the target of malicious backdoor attacks is to mislead the VFL model or harm its overall performance on the original task. Typically, passive parties are the backdoor attackers, while the active party is the victim since only the active party has labels. 

\begin{figure}[!ht]
 \centering
 \includegraphics[width=0.99\linewidth]{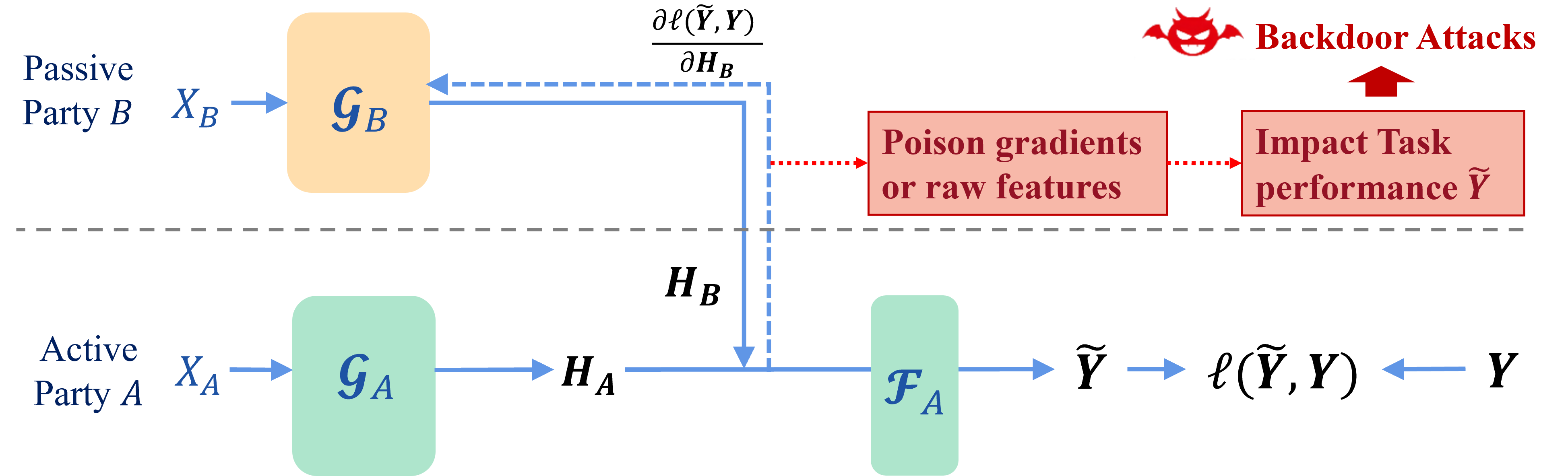}
 \caption{Illustration of backdoor attacks in VFL. Passive parties are the backdoor attackers who aim to impact the task performance of the active party.} 
 \label{fig:sub_fig_vfl_backdoor_attack}
\end{figure}

In the rest of this section, we summarize existing backdoor attacks and defenses. 

\subsubsection{Backdoor Attacks}
Existing research on backdoor attacks can be divided into two main categories, \textit{targeted} and \textit{non-targeted}, depending on whether the attacker has a determinant backdoor target or not. {\blue{Figure \ref{fig:sub_fig_vfl_backdoor_attack} illustrates backdoor attacks}},
and Table \ref{table:backdoor_attacks} summarizes the settings and methods for existing backdoor attacks. 
\begin{table*}[!ht]
	\caption{Summary of existing backdoor attacks in literature.}
	\centering
	\scriptsize
	\begin{tabular}{m{1.7cm}<{\centering}||m{3.5cm}<{\centering}|m{2.7cm}<{\centering}|m{1.6cm}<{\centering}|m{1.2cm}<{\centering}|m{2.0cm}<{\centering}}
	    \hline
		~ & \shortstack{Attacking Method} & 
		\shortstack{VFL \\ Setting} &
		\shortstack{Against \\ Protocol} &
		\shortstack{Attacking \\ Phase} & 
		\shortstack{Auxiliary \\ Requirement} \\
	    \hline
	    \hline
		\multirow{3}*{\shortstack{Targeted\\Backdor \\Attack}} & Label Replacement Backdoor (LRB)\cite{liu2020backdoor} & \multirow{1}*{aggVFL} & \multirow{1}*{P-2} & \multirow{1}*{Training} &  $\geq 1$ label of clean samples \\
		\cline{2-6}
		~ & Adversarial Dominant Input (ADI)\cite{pang2022attacking} &\multirow{1}*{ VLR/splitVFL$_c$} & \multirow{1}*{P-0(g)/P-1} & \multirow{1}*{Inference} & a few samples of other party\\
        \hline
        \multirow{3}*{\shortstack{Non-targeted\\Backdoor \\Attack}} & \multirow{1}*{Adversarial attack\cite{goodfellow2015explaining,liu2021rvfr}} & splitVFL/aggVFL & P-1 & Training  & \textendash  \\
        \cline{2-6}
		~ & \multirow{1}*{Missing attack\cite{liu2021rvfr}} & splitVFL/aggVFL & P-3 & Training  & \textendash \\
	   \cline{2-6}
		~ & \multirow{1}*{Graph-Fraudster\cite{chen2021fraudster}} & splitVFL & P-2 & Inference  & \textendash \\
		\hline
	\end{tabular}
\label{table:backdoor_attacks}
\end{table*}

\begin{table}[!tb]
\centering
\scriptsize
\caption{Summary of defense strategies for defending against backdoor attacks.}
\begin{tabular}{m{1.3cm}<{\centering}||m{1.0cm}<{\centering}|m{5.0cm}<{\centering}|m{3.5cm}<{\centering}}
\hline
\multicolumn{1}{c||}{\shortstack{Defense\\Work}}  &
\multicolumn{1}{c|}{\shortstack{VFL\\Setting}}  &
\multicolumn{1}{c|}{\shortstack{Defense\\Scheme}} &
\multicolumn{1}{c}{\shortstack{Against\\Attack}}\\

\hline
\hline
 DP\cite{zou2022defending} & aggVFL & Add Noise & {\shortstack{Targeted}} \\
\hline
 GS~\cite{zou2022defending} & aggVFL & Sparsify Gradient & {\shortstack{Targeted}}\\
\hline
 CAE~\cite{zou2022defending} & aggVFL & HE+Disguise Label & {\shortstack{Targeted}}\\
\hline
 DCAE~\cite{zou2022defending} & aggVFL  & HE+Disguise Label+DG  & {\shortstack{Targeted}}\\
\hline
 RVFR~\cite{liu2021rvfr} & splitVFL & Robust Feature Sub-space Recovery & {\shortstack{Targeted}/Non-targeted} \\
\hline
\end{tabular}
\label{tab:backdoor_defense_comparison}
\end{table}

\textbf{Targeted backdoor attacks} secretly train a model that achieves high performance on both the original and the targeted backdoor tasks. The objective function of targeted backdoor attacks can be written as follows:
\begin{equation*}\label{eq:targeted_backdoor_objective}
\hspace{-0.2cm} \min_{\Theta} \mathcal{L}_{\text{BD}}(\Theta; \mathcal{D}) \triangleq \frac{1}{N_{cln}}\sum_{i \in \mathcal{D}_{cln}} \ell(\tilde{y}_i,y_i) + 
\frac{1}{N_{poi}}\sum_{i \in \mathcal{D}_{poi}} \ell(\tilde{y}_i,\tau)
\end{equation*}
where $\tilde{y}_i$ is the prediction for sample $x_i$, subscripts $_{cln}$ and $_{poi}$ are short for "clean" and "poisoned" respectively, $\tau$ denotes the target label chosen by the attacker. 

Liu et al.~\cite{liu2020backdoor} proposed a Label Replacement Backdoor attack (LRB), in which the attacker replaces the gradients of a triggered sample with the ones of a clean sample of the targeted class to achieve a high backdoor accuracy while keeping the main task accuracy at a high level. Pang et al.~\cite{pang2022attacking} introduced the Adversarial Dominating Input (ADI), which is an input sample with features that override all other features and lead to a certain model output, and proposed gradient-based methods in both white-boxed and black-boxed settings.

\textbf{Non-targeted backdoor attacks}, similar to Byzantine attacks \cite{alistarh2018byzantine} that are typically studied in HFL, aim to hurt the convergence or the performance of the original task by using adversarial samples~\cite{goodfellow2015explaining,liu2021rvfr}, noisy samples or missing features\cite{liu2021rvfr}.
An adversarial sample is generated using the Fast Gradient Sign Method (FGSM), in which a perturbation $\Delta x_i = \epsilon \text{sign}(\frac{\partial \ell}{\partial x_i})$ is added to the original sample $x_i$ where $\epsilon$ is the magnitude of the perturbation~\cite{goodfellow2015explaining}. Multiple research works~\cite{goodfellow2015explaining,liu2021rvfr} demonstrate the effectiveness of this kind of attack in its misleading performance. If $\Delta x_i$ is simply a randomly generated perturbation, then the attack is referred to as the noisy-sample attack. 

The missing-feature attack simulates real-world VFL scenarios with unstable network\cite{liu2021rvfr} in which, for example, the local model output of a passive party may failed to reach the active party for collaboration. 

\subsubsection{Defense Strategies}
Traditional defense strategies such as adding noise and GS are effective in defending against targeted and non-targeted backdoor attacks\cite{liu2021rvfr,zou2022defending}. However, these defenses suffer from trade-offs between main task accuracy and backdoor task accuracy. On the other hand, cryptographic defense strategies are generally ineffective for defending against backdoor attacks because they preserve the computed outputs and thus do not impact the backdoor training objectives. In \cite{liu2020backdoor}, the authors show that gradient-replacement backdoor attacks can still survive in HE-protected VFL protocols.




Therefore, emerging defense strategies have been proposed to further improve the effectiveness of defenses. For example, CAE and DCAE both show promising effectiveness in defending against targeted backdoor attack~\cite{zou2022defending}. 
RVFR~\cite{liu2021rvfr} is put forward to defend against both target and non-target backdoor attacks in VFL scenarios by robust feature subspace recovery. We compare these defenses in Table \ref{tab:backdoor_defense_comparison}.


In summary, research works on defending backdoor attacks in VFL are still at an early stage. It is worth exploring new effective defense strategies while maintaining good model utility.

\section{Data Valuation and Fairness}
\label{fairness}
VFL opens up new opportunities for cross-institution and cross-industry collaborations.  As industrial use cases grow, a critical challenge for establishing a stable and sustainable federation among parties is the lack of fair data valuation and incentive design to allocate profits. In addition, a responsible VFL framework should also address various bias problems towards certain groups of people. In this section, we discuss the research progress for data valuation, explainability, and fairness for VFL.

\subsection{Data Valuation} 
Currently, most research works on data valuations for FL framework still focus on HFL scenarios \cite{yu2020fairness,Song2019profit, wang2020principled, liu2022gtg}, while data valuations on VFL are much less studied. \cite{ShapleyVFL,Wang2019Shapley} are among the earliest works that proposed contribution evaluation frameworks for VFL using Shapley valuations on features. Shapley-based approaches typically adopt model performance gain as a key metric to measure data value. \cite{FedValueVFL} proposed a model-free approach that uses conditional mutual information for Shapley to evaluate the feature importance and data values in VFL. \cite{fan2022fair} proposed an embedding-based Shapley evaluation method for VFL and applied this method to both asynchronous and synchronous settings. 
{\blue{\cite{NEURIPS2022_vfps} focused on party-level evaluation from a mutual information (MI) perspective and adopted such evaluations to select important participants to improve the scalability of VFL. However, Shapley-based and MI-based evaluations}} are computationally challenging, which makes them difficult to apply to real-world cases. Improving the efficiency of Shapley calculations is an important future research direction. 
 
 \subsection{Explainability}
In fields that are highly regulated, such as financial and medical fields, making the trained VFL model explainable to authorities and compliance is of paramount importance. Currently, only a limited amount of works are proposed to address explainability of VFL. For example, \cite{chen2022evfl} proposed an explainable VFL framework using credibility assessment and counterfactual analysis {\blue{to control data quality and explain counterfactual instances}}. \cite{zheng2020vertical} designed a VFL scheme {\blue{based on logistic regression with bounded constraints}} for interpretable scorecards in credit scoring. \cite{kang2022prada} proposed a feature grouping method that converts original features with low explainability into explainable feature groups to enhance the explainability of VFL prediction models. While designing VFL with explainability is an important research topic, how to reconcile privacy preserving and explainability in VFL is also a crucial research direction because the two objectives may contradict each other. 

\subsection{Fairness}
Machine learning models trained in a collaborative setting may inherit bias towards certain user groups. Addressing fairness problem in VFL is an emerging research topic. FairVFL \cite{qi2022fairvfl} is a framework to use adversarial learning to remove bias for the fairness-sensitive features in a privacy-preserving VFL setting. \cite{liu2021achieving} provided a fairness objective in VFL and developed an asynchronous gradient coordinate-descent ascent algorithm to solve it. The core challenge for addressing fairness in VFL is to identify fairness-sensitive features and perform collaborative debias training while preserving data privacy and protocol efficiency. 

\subsection{Datasets}\label{dataset}

We list datasets commonly used in current VFL works in Table \ref{tab:dataset}. Most of the datasets used in VFL research are tabular datasets from Finance, Healthcare, and Advertising. This manifests that, on the one hand,  VFL has a broad range of applications in the three fields. On the other hand, tabular datasets dominate VFL research for their convenience in forming multi-party scenarios in VFL, indicating that we are short of research datasets of diverse types (e.g., image, text, or video). In addition, only NUSWIDE and Vehicle datasets consist of multi-modal features that can naturally simulate the two-party VFL scenario. Other datasets listed in Table \ref{tab:dataset} are adopted from existing machine learning research works, and there is no established way for VFL researchers to partition these datasets for VFL research. Therefore, facilitating industrial applications and academic research in the VFL area calls for practical datasets and high-quality benchmarks. 

\begin{table*}[ht!]
\centering
\footnotesize
\caption{Commonly used datasets in VFL. In the Size column, the number represents the total amount of samples of each dataset. For the three graph datasets, the number on the left of / represents the number of nodes, while the number on the right represents the number of edges. }
\begin{tabular}{c||c|c|c}
\hline
\\[-1em]
Dataset & Data Type & Size & Description \\
\hline
\hline
 Income~\cite{kohavi1996scaling} & Tabular & 48842 & Demographics and income features \\ 
\hline
\\[-1em]
 Bank~\cite{moro2014data} & Tabular & 41188 & Demographics and economic features \\ 
\hline
\\[-1em]
 Credit Card~\cite{yeh2009comparisons} & Tabular & 30000 & Demographics and payments  \\ 
\hline
\\[-1em]
Give Me Some Credit~\cite{givemesomecredit} & Tabular & 250000 & Debt features  \\
\hline
\\[-1em]
 MIMIC III~\cite{johnson2016mimic} & Tabular & 42276 & Medical records \\
\hline
\\[-1em]
Breast Cancer~\cite{street1993nuclear}  & Tabular & 569& Breast tumor features \\ 
\hline
\\[-1em]
 Diabetes~\cite{smith1988using} & Tabular & 400  & Patient records \\ 
\hline
Avazu~\cite{avazu} & Tabular & 4M &  Click-through data  \\
\hline
\\[-1em]
 Criteo~\cite{criteodata} & Tabular & 4.5M & Click-through data \\
\hline
\\[-1em]
Vehicle~\cite{duarte2004vehicle} & Tabular & 98528 & Acoustic and seismic signals \\ 
\hline
\\[-1em]
 Drive~\cite{Dua2019} & Tabular & 58509 & Electric current drive signals\\
\hline
 Cover type~\cite{blackard1999comparative} & Tabular & 581012 & Digital spatial data \\ 
\hline
\\[-1em]
 NUSWIDE~\cite{nus-wide-civr09} & Tabular & 269648 & Image and the associated tags from Flickr \\
\hline
\\[-1em]
Handwritten~\cite{van1998handwritten} & Tabular & 2000 & Handwritten digit features \\
\hline
\\[-1em]
 Epsilon~\cite{epsilon} & Tabular & 500000 & Synthetic data \\
\hline
\\[-1em]
BHI~\cite{bhi} & Image & 277524 & Medical images \\
\hline
\\[-1em]
CheXpert~\cite{irvin2019chexpert} & Image & 65240  & Medical images \\
\hline
\\[-1em]
Modelnet~\cite{wu20153d} & Image & 20000  & Multi-views of 3D objects  \\
\hline
 Cora~\cite{sen2008collective} & Graph & 2708/5429 &  Citation network \\
\hline
Citeseer~\cite{sen2008collective} & Graph & 3327/4732 & Citation network \\
\hline
 PubMed~\cite{sen2008collective} & Graph  & 19717/44338 & Citation network \\
\hline
Yahoo Answers~\cite{yahoo} & Text & 1.46M & Corpus of questions and answers \\
\hline
\\[-1em]
News20~\cite{keerthi2005modified} & Text & 19928 &  Newsgroup documents \\ 
\hline
\hline
\end{tabular}
\label{tab:dataset}
\end{table*}

\section{VFLow: A VFL Optimization Framework}\label{vflow}

\begin{figure}[!ht]
\centering
\includegraphics[width=1.0\linewidth]{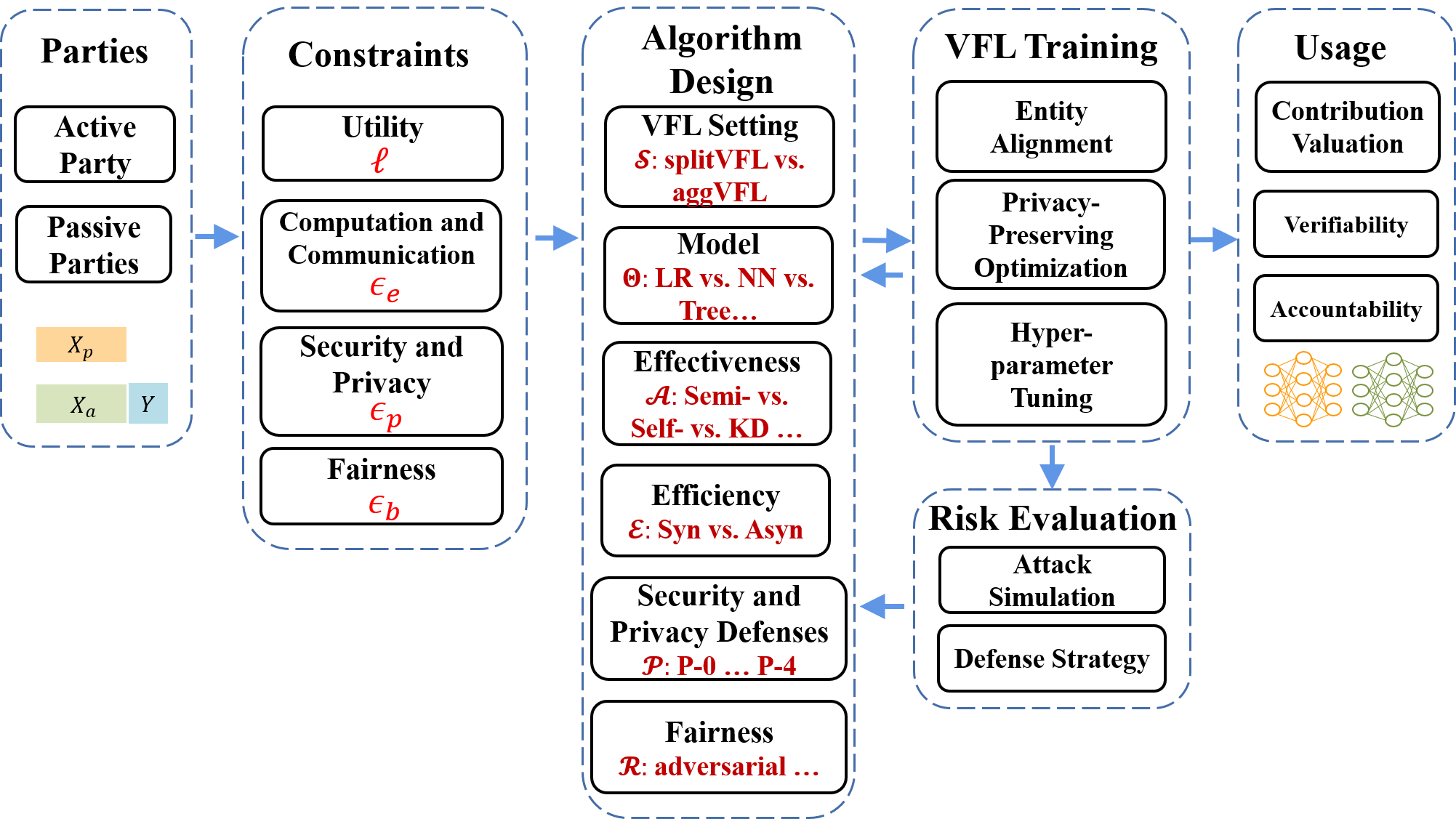}
\caption{VFLow: A Framework for setting up, designing and optimizing VFL algorithms.}
\label{fig:meta-objective}
\end{figure}

We propose a comprehensive VFL optimization framework consisting of major considerations for setting up and optimizing a VFL algorithm, as illustrated in Figure \ref{fig:meta-objective}. We termed this framework \textbf{VFLow}. 

In VFLow, we take into account major constraints, including privacy, efficiency, and fairness, to guide the design of a VFL algorithm from aspects of the model architecture and partition settings, effectiveness and efficiency improving strategies, privacy defense strategies, as well as fairness improving strategies covered in this work. In addition, VFLow consists of a separate risk evaluation module that comprehensively evaluates data attacks and defense strategies. Finally, for model usage, party contributions, accountability, and verifiability tools are necessary for a sustainable and trustworthy federation (also see Sec. \ref{challenge}). We further extend the objective function formulated in Eq. \ref{eq:problem} to a more general meta-objective, in which we want to minimize the main task loss (i.e., maximize utility) constrained by privacy, efficiency (i.e., communication and computation), and fairness:
\begin{align}\label{eq:vflow_2}
\begin{split}
\hspace{-0.2cm} &  \min_{\Theta} \ell(\Theta;\mathcal{S},\mathcal{A},\mathcal{E},\mathcal{P},\mathcal{R},\mathcal{D}) \\ &
s.t. \ \ M_{p}(\Theta;\mathcal{K},\mathcal{P}) \le \epsilon_p, M_{e}(\Theta;\mathcal{E},\mathcal{P})) \le \epsilon_e, M_{b}(\mathcal{R},\mathcal{D}) \le \epsilon_b
\end{split}
\end{align}
where $\Theta$ and $\mathcal{S}$ denote specific models and a VFL setting, respectively; $\mathcal{A}$ denotes an effectiveness improving strategy, $\mathcal{P}$ denotes a privacy defense strategy, $\mathcal{K}$ denotes the collection of attack algorithms, $\mathcal{E}$ denotes an efficiency improving strategy, and $\mathcal{R}$ denotes a fairness improving strategy. $M_p$ denotes a measurement for measuring privacy leakage imposed by attacks $\mathcal{K}$ against the defense strategy $\mathcal{P}$. $M_e$ is the efficiency measure, typically with respect to communication load and computation resources. $M_b$ measures the system bias. $\epsilon_p$, $\epsilon_e$, and $\epsilon_b$ are constraints for privacy leakage, efficiency cost, and bias, respectively.

{\blue{This optimization problem can be considered as a constrained multi-objective federated learning problem ~\cite{kang2023optimizing}. Such formulation brings about a set of solutions, each of which is an optimal trade-off between multiple objectives and thus provides stakeholders with flexible decision options.
}
}

\section{Applications}
\label{applied}
Due to its practical merits for enabling data collaboration between multiple institutions across industries, VFL has attracted increasing attention from both academia and industry. In this section, we provide an overview of VFL applications.

\textbf{Recommendation systems} are typically adopted in VFL to support advertising applications. {\blue{Federated bandit can be used as a promising technique ~\cite{LiuRecBandits2022,Libandits2022,ZhuFedBandit2021} for FL.}} Shmueli et al.~\cite{Erez2017PPCF} proposed a privacy-preserving collaborative filtering protocol. Atarashi et al.~\cite{Kyohei2021HOFM} proposed a higher-order factorization machine in the VFL setting. Recommendation systems can be built between two platforms holding different rating data. Cui et al.~\cite{cui2021exploiting} proposed a secure cross-platform recommendation based on secure computation protocols. Zhang et al.\cite{zhang2021vertical} proposed a VFL recommendation based on clustering and latent factor model to reduce the dimension of the matrix and improve the recommendation accuracy. To achieve privacy-preserving recommendations based on the personal data cloud, Yuan et al.\cite{yuan2022privacy} proposed a hybrid federated learning recommendation algorithm named HyFL, which 
exploits the advantages of both HFL and VFL. Cai et al.~\cite{cai2022socialrec} proposed a DP-based VFL recommendation framework between a social recommender system and a user social graph. 

Many internet companies have adopted VFL to support advertising. For example, ByteDance developed a tree-based VFL algorithm based on the Fedlearner framework, which significantly improves its advertising efficiency \cite{cai2020Bytedance}. Based on the VFL module in its 9N-FL framework, JD has established a joint model for advertising, which has promoted the cumulative increase of all participating parties' income \cite{Hou2021JD}. Tencent applied its Angel PowerFL platform to establish a VFL federation between advertisers and advertising platforms to boost model accuracy\cite{tencent2021}.
Based on the trusted intelligent computing service framework (TICS), Huawei applied VFL to advertising\cite{Wu2022Huawei} to leverage user profile and behavior data dispersed in different platforms.



\textbf{Finance} is another major application that new VFL approaches have been rapidly developed. For example, a gradient-based method for traditional scorecard model training is proposed in~\cite{zheng2020vertical}. In \cite{chen2021homomorphic}, a secure large-scale sparse logistic regression algorithm is designed and applied to financial risk control. Kang et al.~\cite{kang2022prada} developed a fine-grained adversarial domain adaptation algorithm to address the label deficiency issue in the financial field. Long et al.~\cite{long2020openbanking} discussed the applications and open challenges for FL in open banking. Wang et al.~\cite{Wang2019Shapley} provided an overview of the use cases of FL in the insurance industry. WeBank uses customers' credit data and invoice information from partner companies to jointly build a risk control VFL model \cite{Cheng2020CACM}. 



\textbf{Healthcare}  has been very active in applied research in VFL. A privacy-preserving logistic regression is proposed in \cite{hu2022vertical} and applied to clinical diagnosis. Chen et al.\cite{chen2020vafl} proposed an asynchronous VFL framework and verified the effectiveness of this framework on the public health care dataset MIMIC-III. In~\cite{rooijakkers2020convinced}, the authors applied VFL to cancer survival analysis to predict the likelihood of patients surviving time after diagnosis and to analyze which features might be associated with the chance of survival. \cite{cha2021implementing} proposed an efficient VFL method using autoencoders to predict hearing impairment after surgery based on a vestibular schwannoma dataset.
Song et al.\cite{song2021federated} applied VFL to the joint modeling between mobile network operators (MNOs) and health care providers (HP). 

\textbf{Emerging applications} have also been exploited in recent years for discovering novel data utilization in fields such as electric vehicles and wireless communications. Teimoori et al.\cite{teimoori2022secure} proposed a VFL algorithm to locate charging stations for electric vehicles while protecting user privacy. \cite{Solmaz2020wireless} discussed the opportunities for VFL to be utilized in 5G wireless networks. \cite{zhang2020radionetworks} proposed a VFL-based cooperative sensing scheme for cognitive radio networks. \cite{hashemi2021optical} developed a VFL framework for optical network disaggregation. \cite{liu2021smartgrid} applied VFL to collaborative power consumption predictions in smart grid applications. \cite{ge2021failureprediction} proposed VFL modelings for predicting failures in intelligent manufacturing. 

\textbf{MultiModal Tasks} are performed when participants in VFL hold data from multiple modalities, such as vision, language, and sense. Liu et al.~\cite{liu2020vision-language} proposed an aimNet that helps the FL model learn better representations from textual and visual features through multi-task learning. 
Liang et al.~\cite{liang2021selfsupervised} proposed a self-supervised vertical federated neural architecture search approach that automatically optimizes each party's local model for the best performance of the VFL model, given that participating parties hold heterogeneous image data. \textbf{Vertical federated graph learning (VFGL)} algorithms are proposed to leverage features, relations, and labels that belong to the same group of people but are dispersed among different organizations. VFGNN~\cite{chen2022verticalgraph} and FedVGCN~\cite{ni2021fedvgcn} perform node classification on the scenario where all parties share the same set of nodes, but each party only owns partial features and relations of these nodes. FedSGC~\cite{cheung2021Fedsgc} performs node classification on another scenario where one party has only graph structural information while other parties have only node features.

\section{Open Challenges and Future Direction}
\label{challenge}
In this section, we discuss some of the major open challenges facing the development of VFL frameworks and propose possible paths in the future. 

\textbf{Interoperability.} Thanks to the rapid development of efficient privacy-preserving technologies in recent years, more and more VFL projects and open-sourced platforms have been developed and applied in real-world scenarios, connecting data silos in various industries. However, the lack of interoperability of existing frameworks has become a new pain point for its industrial growth. Different platforms adopt different sets of secure computation and privacy-preserving training protocols, making cross-platform collaboration difficult and turning data silos into platform silos. One possible route to solve this challenge is to enforce the interoperability of platforms by developing algorithm and architecture standards so that platforms can connect with others more readily. Another route is to develop seed projects to support fundamental functionalities and modules for interoperability as a plug-in tool for diverse platforms.

\textbf{Trustworthy VFL.} To be trustworthy, VFL frameworks must appropriately reflect characteristics such as privacy and security, effectiveness, efficiency, fairness, explainability, robustness, and verifiability. Data needs to be protected in transit and at rest with clear security and privacy definitions and scopes. Despite recent research efforts on this subject, there is still a lack of universally effective defense strategies that are lossless and highly efficient. The trade-off between utility-privacy-efficiency~\cite{zhang2022tradeoff} is still the focus of future studies. Applying multi-objective optimization techniques~\cite{kang2023optimizing} in VFLow is a promising research direction towards trustworthy VFL~\cite{ren2023secureboost}. In addition, the path toward a trustworthy FL framework is for the trained models to be verifiable and auditable. One possible route is for the released trained models in VFL to be protected by verifiable intellectual property (IP) protection methods~\cite{li2022fedipr} in an efficient manner to prevent malicious IP attacks while fulfilling privacy requirements. {\blue{Blockchain is leveraged to address the issue that a vanilla FL framework heavily relies on a central server, which means the system is vulnerable to this party's mal-behavior. How to integrate Blockchain into VFL frameworks to improve the overall security and robustness is an interesting future direction.}}


\textbf{Automated and Blockchained VFL.} Automated machine learning (AutoML) is of great interest in alleviating human effort and achieving satisfactory model performance \cite{yao2018taking}. 
Liang et al.~\cite{liang2021selfsupervised} proposed a Vertical Federated Neural Architecture Search that learns individual model architecture for each client. 
\cite{zhu2021federatedNASbook} discussed challenges in applying NAS to VFL under encryption. For VFL, participants without labels can not perform individual training or evaluation locally. Thus, their hyperparameters are nested in the collaborative training. This unique setting makes AutoML in VFL more challenging. Blockchain is leveraged to address the issue that a vanilla FL framework heavily relies on a central server, which may lead to a single point of failure or privacy vulnerabilities. By utilizing Blockchain, participating parties can exchange their model updates in a decentralized and verifiable manner. 
How to integrate Blockchain into VFL frameworks to improve the overall security and robustness is an interesting future direction.


\section{Concluding Remarks}
Vertical federated learning enables collaborative learning of feature-partitioned data distributed across multiple institutions. It has become an attractive solution for solving industrial data silo problems caused by the enforcement of strict data regulations. Despite its practical usefulness, as evidenced by a growing number of VFL projects and use cases, the breadth and depth of the research advances still lag behind those of HFL. We present an extensive categorization of research efforts and new challenges in VFL and propose a novel framework towards comprehensively formulating relevant aspects of VFL. We hope this work will encourage future research efforts to address these challenges in this area.

\bibliographystyle{IEEEtran}
\bibliography{ref_yang,ref_zou}

\end{document}